\documentclass[journal]{IEEEtran}
\usepackage{algpseudocode}
\usepackage{algorithm}
\usepackage{amsmath}
\usepackage{cite}
\usepackage{subfloat}
\usepackage{hhline}
\usepackage{multirow}
\usepackage{amsfonts}
\usepackage{array}
\usepackage{graphicx}
\usepackage{epstopdf}
\usepackage{balance}
\usepackage{tabularx}
\usepackage[hyphens]{url}

\ifCLASSOPTIONcompsoc
\usepackage[caption=false, font=normalsize, labelfont=sf, textfont=sf]{subfig}
\else
\usepackage[caption=false, font=footnotesize]{subfig}
\fi

\newcommand\blfootnote[1]{%
	\begingroup
	\renewcommand\thefootnote{}\footnote{#1}%
	\addtocounter{footnote}{-1}%
	\endgroup
}
\begin{document}

\title{Superpixel Based Graph Laplacian Regularization for Sparse Hyperspectral Unmixing}
\author{Taner~Ince,~\IEEEmembership{Member,~IEEE}
\thanks{The author is with the Department of Electrical and Electronics Engineering,
	Gaziantep University, 27310 Gaziantep, Turkey (e-mail:
	tanerince@gantep.edu.tr}}

\markboth{IEEE GEOSCIENCE AND REMOTE SENSING LETTERS, VOL. ~XX, NO. ~X, July~2020}%
{Shell \MakeLowercase{\textit{et al.}}: Bare Demo of IEEEtran.cls for IEEE Journals}

\maketitle

\begin{abstract}
An efficient spatial regularization method using superpixel segmentation and graph Laplacian regularization is proposed for sparse hyperspectral unmixing method. Since it is likely to find spectrally similar pixels in a homogeneous region, we use a superpixel segmentation algorithm to extract the homogeneous regions by considering the image boundaries. We first extract the homogeneous regions, which are called superpixels, then a weighted graph in each superpixel is constructed by selecting $K$-nearest pixels in each superpixel. Each node in the graph represents the spectrum of a pixel and edges connect the similar pixels inside the superpixel. The spatial similarity is investigated using graph Laplacian regularization. Sparsity regularization for abundance matrix is provided using a weighted sparsity promoting norm. Experimental results on simulated and real data sets show the superiority of the proposed algorithm over the well-known algorithms in the literature.
\end{abstract}


\IEEEpeerreviewmaketitle

\section{Introduction}
\blfootnote{This work has been submitted to the IEEE for possible publication. Copyright may be transferred without notice, after which this version may no longer be accessible.}
\IEEEPARstart{H}{yperspectral} imaging used in remote sensing allows the identification of the substances in the scene by measuring the light spectrum over hundreds of contiguous bands. However, low spatial resolution of hyperspectral sensor and combination of different materials in the homogeneous mixtures cause mixed pixels. Decomposition of a mixed pixel into spectral signatures (endmembers) with corresponding fractions (abundances) is known as spectral unmixing \cite{974727}.

Spectral unmixing methods mainly use linear mixture model (LMM) in which the observed spectra is a linear combination of the endmembers with corresponding abundances. Linear spectral unmixing (LSU) methods are simple and has tractable solutions; however, nonlinearity and spectral variability effects the performance of spectral unmixing \cite{8528557}. In LSU, an endmember extraction step is applied and then abundance value for each pixel is estimated. There are many algorithms for endmember extraction such as N-FINDR \cite{366289}, pixel purity index (PPI) \cite{Boardman-etal1995} and vertex component analysis (VCA) \cite{1411995}. These algorithms require pure pixel assumption and it is not always satisfied due to the spatial resolution. One way of solving this problem is to use ground spectral libraries and then obtaining the abundance value of each pixel using this large spectral library. Generally, the number of endmembers in the scene are small compared to the number of endmembers in the spectral library. This means that only small number of endmembers contribute the mixed pixel. Therefore, abundance vector of mixed pixel is expected to be sparse. Estimating the sparse abundance vector using \emph{a priori} available spectral library is known as sparse unmixing (SU) \cite{5594963}. Sparse unmixing by variable splitting and augmented Lagrangian (SUnSAL) \cite{5594963} solves an $l_{1}$ minimization problem satisfying abundance non-negativity constraint (ANC) and abundance sum constraint (ASC). Collaborative SUnSAL (CLSUnSAL) \cite{6471206} solves an $l_{2,1}$ norm optimization problem to promote the row-sparse structure of the abundance matrix. Local collaborative sparse unmixing (LCSU) \cite{7433396} estimates the sparse abundance matrix by solving CLSUnSAL in a neighborhood of pixels to obtain more accurate abundance values. Iterative reweighted sparse unmixing (IRWSU) \cite{7210209} use a weighting strategy in the formulation which has a better abundance estimation compared to CLSUnSAL. Furthermore, nonconvex sparsity based methods are developed for hyperspectral unmixing \cite{8616834}.

Furthermore, spatial-contextual information of the abundance map is exploited in many works \cite{6196219,8360169,7935525,8288814} by considering the piecewise smoothness of the abundance map. Total variation (TV) regularization \cite{Rudin} is used in SUnSAL-TV \cite{6196219} which minimizes the fractional abundance of neighboring pixels. It provides smooth abundance map, however it does not take into account the local changes in the abundance map. A spatial discontinuity weight strategy is developed to preserve the details in the abundance map better \cite{8360169} using the idea that smooth abundance map condition is not hold in real scenarios. A double reweighted sparse unmixing and TV (DRSU-TV) \cite{7935525} improves the sparsity of the abundance matrix by using a double reweighting strategy which is applied in both spectral and spatial domains. Spectral-spatial weighted sparse unmixing ($\textmd{S}^{2}\textmd{WSU}$) method is presented in \cite{8288814} which includes a single regularizer with spectral and spatial weighting matrices in the proposed formulation to improve the abundance estimation.

It is known that similar pixels in a local region are likely to have similar abundances, graph based approaches are developed for hyperspectral unmixing \cite{7373556,7178248,8725887}. A hypergraph-regularized sparse nonnegative matrix factorization (NMF) \cite{7373556} based unmixing approach employs a hypergraph structure where each pixel is taken as a vertex and the pixels in the neighborhood of that pixel form a hypergraph. Therefore, similar pixels having similar abundances are found which leads to obtain better unmixing results. In a same manner, graph Laplacian regularization is used in \cite{7178248} to promote the smoothness of the abundance map in sparse regression framework. Recently, spatial-contextual information is exploited using hypergraph learning \cite{8725887} to extract the similarity between the pixels in a small spatial neighborhood. \cite{8725887} uses $K$-nearest-neighbors algorithm to find the spectrally similar pixels in a local neighborhood. However, as the noise level increases it is difficult to find the spectrally similar pixels in a local region. Furthermore, it is likely to find similar pixels with different regions of data separated by edges in the image.

Recently, superpixel segmentation is investigated in several works \cite{7042923,8486711,8900376,8536490,9099278} in hyperspectral imaging. Fang \textit{et. al} \cite{7042923} propose a hyperspectral classification method based on superpixel segmentation. The pixels in each superpixel are jointly represented by a set of common atoms. \cite{8486711} employs multiscale superpixels to extract the local information for hyperspectral image (HSI) classification. A superpixel weighting strategy is used in \cite{8900376} to include the spatial correlation. A fast multiscale spatial regularization based on simple linear iterative clustering (SLIC) \cite{6205760} is proposed in \cite{8536490}. Superpixel-based reweighted low-rank and total variation (SUSRLR-TV) \cite{9099278} minimizes the rank of the abundance matrix in each superpixel and promote the smoothness of the abundance map using TV.

In this paper, we propose a superpixel based graph Laplacian for sparse unmixing (SBGLSU). Superpixel segmentation takes into account the image boundaries when segmenting the HSI into homogeneous regions. Therefore, we first segment HSI into many superpixels using SLIC. Then, a weighted graph for each superpixel is constructed, where each node represents the neighboring pixels in the superpixel. Although, superpixels are homogeneous shape adaptive spatial-neighboring pixels, we include a weighted graph regularization to measure the similarity of the $K$-nearest pixels in each superpixel. In this manner, spatial correlation among the $K$-nearest neighboring pixels inside the superpixel is better extracted. The sparsity of the abundance matrix is satisfied using an $l_{1}$ norm regularizer with a weighting strategy that promotes the joint-sparsity of the abundance matrix.

The rest of the paper is organized as follows. Section II explains the proposed method. The simulated and real data experiments are given in Section III. Finally, Section IV concludes the paper.
\section{Superpixel Based Graph Laplacian Regularization for sparse unmixing (SBGLSU)}
LMM assumes that endmembers are linearly combined to form the measured spectrum of a pixel. It can be modeled as
\begin{equation}\label{eq1}
\nonumber
\mathbf{Y} = \mathbf{AS}+\mathbf{N}
\end{equation}
where $\mathbf{Y} = [\mathbf{y}_{1},\ldots,\mathbf{y}_{n}]\in \mathbb{R}^{L\times n}$ is the $L$-band spectrum of $n$ pixels where each $\mathbf{y}_{i}$ $(i=1,2,\ldots,n)$ represents the spectrum of $i$th pixel in the HSI, $\mathbf{A}\in \mathbb{R}^{L\times m}$ is the mixing matrix containing $m$ endmembers, $\mathbf{S} = [\mathbf{s}_{1},\ldots,\mathbf{s}_{n}]\in \mathbb{R}^{m\times n}$ fractional abundance matrix where each $\mathbf{s}_{i}$ $(i=1,2,\ldots,n)$ represents the fractional abundance vector of $i$th pixel and $\mathbf{N} \in \mathbb{R}^{L\times n}$ models the error in the measurements. If the number of active endmembers is much lower than the number of endmembers in spectral library $\mathbf{A}$, then abundance matrix $\mathbf{S}$ is expected to be sparse.

Furthermore, a spatial similarity exists between neighboring pixels in a HSI which leads to abundance similarity of neighboring pixels. Therefore, we first construct a weighted graph $G=(V,E)$ \cite{von2007tutorial} where $V=\{{v}_{1},\ldots,{v}_{n}\}$ and $E=\{{e}_{1},\ldots,{e}_{n}\}$ denote the vertex set and weighted edge set, respectively. A weighted adjacency matrix $W\in \mathbb{R}^{n\times n}$ is constructed where each entry $W_{ij}$ defines the degree of similarity between the spectrum of pixels $\mathbf{y}_{i}$ and $\mathbf{y}_{j}$. If $\mathbf{y}_{i}$ and $\mathbf{y}_{j}$ are similar then a large positive weight is assigned to $W_{ij}$. If they are not similar, a small positive value is assigned. There are different choice of selecting adjacency matrix to construct similarity graphs. We use Gaussian heat kernel which is defined as
\begin{equation}\label{gr1}
    W_{ij}=\textmd{exp}\bigg(-\frac{\|\mathbf{y}_{i}-\mathbf{y}_{j}\|_{2}^{2}}{2\sigma^{2}}\bigg)
\end{equation}
where $\sigma$ controls the width of neighborhood.

When constructing a similarity graph for hyperspectral data using weighted adjacency matrix, it is likely to find two similar pixels in different regions of the HSI. However, local regions tend to have similar pixels leading to similar abundances. Therefore, we extract the homogeneous regions using a segmentation algorithm. Generally, $K$-means algorithm is used to extract the local regions, however $K$-means search the whole image to find the similar pixels. For this reason, our purpose is to search a local area which have spatially similar regions. We resort the SLIC to segment the HSI into homogeneous regions. SLIC is a variant of $K$-means clustering but it searches a limited region and it takes into account the image boundaries. It is also easy to use, fast and memory efficient and it requires little number of parameters.

Therefore, we first segment the HSI into superpixels and then construct graph Laplacian for each superpixel. We can express the abundance similarity in all superpixels as
\begin{equation}\label{gr2}
    \frac{1}{2}\sum_{g=1}^{n_{g}}\sum_{(i,j)\in\varepsilon_{g}}W_{g_{ij}}\|\mathbf{s}_{i}-\mathbf{s}_{j}\|_{2}^{2} =\sum_{g=1}^{n_{g}} \textmd{Tr}(\mathbf{S}_{g}\mathbf{L}_{g}\mathbf{S}_{g}^{T})
\end{equation}
Here, $n_{g}$ denotes the number of superpixels in the image, $\varepsilon_{g}$ is the neighborhood of each superpixel, $\textmd{Tr}(\cdot)$ denotes the trace of a matrix, $\mathbf{S}_{g}$ is the abundance matrix of the $g$th superpixel, $\mathbf{L}_{g}=\mathbf{D}_{g}-\mathbf{W}_{g}$ is the graph Laplacian matrix of $g$th superpixel where $\mathbf{W}_{g}$ is the adjacency matrix of $g$th superpixel, $\mathbf{D}_{g}$ is a diagonal matrix which is calculated as $\mathbf{D}_{g_{ii}}=\sum_{j=1}^{n}W_{g_{ij}}$ where $W_{g_{ij}}$ denotes the each entry of $\mathbf{W}_{g}$ .

After defining the abundance similarity measure in each superpixel, SBGLSU is proposed as
\begin{align}\label{gr4}
\nonumber
&\min_{\mathbf {S}} \frac {1}{2} \| \mathbf {Y}- \mathbf {A} \mathbf {S} \|_{F}^{2} + \lambda_{s} \|\mathbf{W}_{s}\odot \mathbf {S} \|_{1}+\lambda_{g}\sum_{g=1}^{n_{g}}\textmd{Tr}(\mathbf{S}_{g}\mathbf{L}_{g}\mathbf{S}_{g}^{T})\\
&+\iota_{R_{+}}(\mathbf{S})
\end{align}
where $\|\cdot\|_{F}$ and $\|\cdot\|_{1}$ denote the Frobenius norm and $l_{1}$ norm, respectively. $\lambda_{s}$ and $\lambda_{g}$ are regularization parameters, $\mathbf{W}_{s}$ is the weight matrix to promote the sparsity of $\mathbf{S}$, $\odot$ denotes Hadamard product. $\iota_{R+}(\mathbf{S})$ is indicator function that is equal to zero if $\mathbf{s}\geq0$ and $+\infty$ otherwise.

We split the optimization problem into subproblems using alternating direction method of multipliers (ADMM) \cite{MAL-016} to solve alternately. The optimization problem (\ref{gr4}) can be written in a compact form as
\begin{equation}\label{gr5}
\min_{\mathbf{S,V}}g\mathbf{(V)}\quad \textmd{subject to}\quad \mathbf{GS + BV =0}
\end{equation}
where
\begin{align}\label{gr4}
\nonumber
&g\mathbf{(V)} = \frac{1}{2}\|\mathbf{Y}-\mathbf{V}_{1}\|_{F}^{2} +\lambda_{s} \|\mathbf{W}_{s}\odot\mathbf{V}_{2} \|_{1} \\
&+\lambda_{g}\sum_{g=1}^{n_{g}}\textmd{Tr}(\mathbf {{V}_{3}}_{g}\mathbf{L}_{g}\mathbf {{V}_{3}}_{g}^{T})+\iota_{R_{+}}(\mathbf{V}_{4})
\end{align}
$\mathbf{V}=(\mathbf{V}_{1},\mathbf{V}_{2},\mathbf{V}_{3},\mathbf{V}_{4})$, $\mathbf{G}=[\mathbf{A},\mathbf{I},\mathbf{I},\mathbf{I}]^{T}$ and $\mathbf{B}=\textmd{diag}[-\mathbf{I}]$.

The augmented lagrangian formulation of (\ref{gr5}) is
\begin{equation}\label{gr6}
    \mathcal{L}\mathbf{(V,S,\Lambda)} = g\mathbf{(V)} + \frac{\mu}{2}\|\mathbf{GS + BV - \Lambda }\|_{F}^{2}
\end{equation}
where $\mu>0$ is a penalty parameter and $\mathbf{\Lambda}/\mu$ denotes the Lagrange multipliers.

The algorithm of SBGLSU is shown in Algorithm \ref{alg:SBGLSU}. SBGLSU includes a weighting strategy to promote the row-sparsity of the abundance matrix. However, ADMM requires that all functions should be closed, proper and convex to guarantee convergence. Therefore, we use inner and outer loops in Algorithm \ref{alg:SBGLSU} to make the convergency of the algorithm better. In simulation section, the maximum iteration number of outer and inner loops are set to $l=60$ and $t=8$, respectively.
\begin{algorithm}[!ht]
	\caption{Pseudocode of the proposed SBGLSU}
	\label{alg:SBGLSU}
	\begin{algorithmic}[1]
		\Require
		\textbf{Y}, \textbf{A}, $\lambda_{s}$, $\lambda_{g}$, $\mu>0$, $\epsilon$, SLIC parameters  \vspace{0.1em}
		\Ensure
		 $l=0$, $t = 0$, $\mathbf{S}^{(0)}$, $\mathbf{V}_{1}^{(0)}$, $\mathbf{V}_{2}^{(0)}$, $\mathbf{V}_{3}^{(0)}$, $\mathbf{V}_{4}^{(0)}$ $\mathbf{\Lambda}_{1}^{(0)}$, $\mathbf{\Lambda}_{2}^{(0)}$, $\mathbf{\Lambda}_{3}^{(0)}$, $\mathbf{\Lambda}_{4}^{(0)}$
        \State \hspace{-0.1em}\textbf{for} $g=1$ \textmd{to} $n_{g}$
        \State \hspace{0.5em}$ \mathbf{L}_{{g}}=\mathbf{D}_{g}-\mathbf{W}_{g}$
        \State \hspace{-0.1em}\textbf{end for}

		\State \textbf{repeat} \vspace{0.1em}
          \begin{align*}
         &\hspace {-3em}\mathbf{W}_{s}(:,i)^{(l)}=
           \bigg[\frac{1}{\|(\mathbf{S}^{(l)}-\mathbf{\Lambda}_{2}^{(l)})(1,:)\|_{2}+ \epsilon};\vdots\\&;\frac{1}{\|(\mathbf{S}^{(l)}-\mathbf{\Lambda}_{2}^{(l)})(m,:)\|_{2}+\epsilon}\bigg]\quad i=1,2,\ldots,n
        \end{align*}
        \Repeat \vspace{0.1em}
		\State \hspace{-1.5em}$\mathbf{S}^{(t+1)}={(\mathbf{A}^{T}\mathbf{A} + 3\mathbf{I})}^{-1} \Big[\mathbf{A}^{T}(\mathbf{V}_{1}^{(t)}+\mathbf{\Lambda}_{1}^{(t)}) $
		\Statex \hspace{3em}$+(\mathbf{V}_{2}^{(t)}+\mathbf{\Lambda}_{2}^{(t)})+ (\mathbf{V}_{3}^{(t)}+\mathbf{\Lambda}_{3}^{(t)})+(\mathbf{V}_{4}^{(t)}+\mathbf{\Lambda}_{4}^{(t)})\Big]$
        \State \hspace{-1.5em}$\mathbf{V}_{1}^{(t+1)}= \frac{1}{1+\mu}(\mathbf{Y}+\mu (\mathbf{A}\mathbf{S}^{(t+1)}-\mathbf{\Lambda}_{1}^{(t)}))$
        \State \hspace{-1.5em}$\mathbf{V}_{2}^{(t+1)}= \textmd{soft}(\mathbf{S}^{(t+1)}-\mathbf{\Lambda}_{2}^{(t)},(\lambda_{s}/\mu){\mathbf{W}_{s}}^{(l)})$
		\State \hspace{-1.5em}\textbf{for} $g=1$ \textmd{to} $n_{g}$\vspace{0.1em}
        \State \hspace{-1em}$ \mathbf{V}_{3_{g}}^{(t+1)}=\mu(\mathbf{S}_{g}^{(t+1)}-\mathbf{\Lambda}_{3_{g}}^{(t)})(2\lambda_{g}\mathbf{L}_{g}+\mu\mathbf{I})^{-1}$\vspace{0.2em}
        \State \hspace{-1.5em}\textbf{end for}\vspace{0.1em}
        \State \hspace{-1.5em}$\mathbf {V}_{4}^{(t+1)} = \max(\mathbf {S}^{(t+1)}-\mathbf{\Lambda}_{4}^{(t)}, {\bf {0}})$\vspace{0.1em}
		\State \hspace{-1.5em}$\mathbf{\Lambda}_{1}^{(t+1)}= \mathbf{\Lambda}_{1}^{(t)} - \mathbf{A}\mathbf{S}^{(t+1)} + \mathbf{V}_{1}^{(t+1)}$ \vspace{0.1em}
        \State \hspace{-1.5em}$\mathbf{\Lambda}_{2}^{(t+1)}= \mathbf{\Lambda}_{2}^{(t)} - \mathbf{S}^{(t+1)} + \mathbf{V}_{2}^{(t+1)}$\vspace{0.1em}
        \State \hspace{-1.5em}$\mathbf{\Lambda}_{3}^{(t+1)}= \mathbf{\Lambda}_{3}^{(t)} - \mathbf{S}^{(t+1)} + \mathbf{V}_{3}^{(t+1)}$\vspace{0.1em}
        \State \hspace{-1.5em}$\mathbf{\Lambda}_{4}^{(t+1)}= \mathbf{\Lambda}_{4}^{(t)} - \mathbf{S}^{(t+1)} + \mathbf{V}_{4}^{(t+1)}$\vspace{0.1em}
		\State \hspace{-1.5em}\textbf{Update iteration:} $t\leftarrow t+1$ \vspace{0.1em}
        \State \hspace{-1.5em}$\mathbf{S}^{(l+1)}\leftarrow \mathbf{S}^{(t+1)}$
        \State \hspace{-1.5em}$\mathbf{\Lambda}_{2}^{(l+1)}\leftarrow \mathbf{\Lambda}_{2}^{(t+1)}$
        \State\hspace{-1.5em}\textbf{Update iteration:} $l\leftarrow l+1$ \vspace{0.1em}
        \Until some stopping criteria is satisfied.
	\end{algorithmic}
\end{algorithm}

For the complexity analysis, the most computationally expensive parts are step 6 and step 10 of Algorithm 1. In Step 6, the term $(\mathbf{A}^{T}\mathbf{A} + 3\mathbf{I})^{-1}$ is fixed and can be precomputed to reduce the complexity. Therefore, calculation of $\mathbf{S}^{(t+1)}$ has a computational complexity of $\mathcal{O}(mnL)$. Similarly, the term $(2\lambda_{g}\mathbf{L}_{g}+\mu\mathbf{I})^{-1}$ in step 10 can be precomputed so that updating $\mathbf{V}_{3}^{(t+1)}$ has a computational complexity of $\mathcal{O}(mn_{g}|n_{s}|^2)$ where $|n_{s}|$ denotes the number of pixels in each superpixel. Other terms have computational complexity of $\mathcal{O}(n)$. Therefore, the overall complexity of SBGLSU is $\mathcal{O}(mnL)+\mathcal{O}(mn_{g}|n_{s}|^2)+\mathcal{O}(n)$.
\section{Simulated and Real Data Experiments}
In this section, we demonstrate the performance of the proposed method by using synthetic hyperspectral data sets. We perform two synthetic data experiment to demonstrate the effectiveness of SBGLSU. Signal to reconstruction error (SRE) is used to measure the quality of the unmixing results. It is defined as $\textmd{SRE} = 10\log_{10}(\|\mathbf{S}\|_{F}^{2}/\|\mathbf{S}-\mathbf{\hat{S}}\|_{F}^{2})$ where $\mathbf{S}$ represents the ground truth abundance map and $\mathbf{\hat{S}}$ is the estimated abundance map.
\subsection{Simulated Data Sets}
In the synthetic data experiments, we create a spectral library $\mathbf{A}$ by selecting 240 signatures randomly from digital spectral library (splib06)\cite{vn4312481} obtained from the U.S. Geological Survey (USGS) which contains the spectra of 498 materials measured in 224 spectral bands distributed uniformly in the interval 0.4 and 2.5 $\mu$m. We generate two simulated data sets satisfying ASC and ANC. Simulated data cube 1 (DC1) is created by selecting five spectral signatures randomly from library $\mathbf{A}$ as active endmembers and using the corresponding fractional abundance maps having size of $75\times 75$. For simulated data cube 2 (DC2), we select nine spectral signatures randomly from $\mathbf{A}$ as active endmembers and using the corresponding fractional abundance maps having size of $100\times 100$. Simulated data sets DC1 and DC2 is then contaminated with Gaussian noise of signal-to-noise ratio (SNR) with SNR=20, 30 and 40 dB, respectively.

\subsection{Comparison to Other Unmixing Methods}
\begin{figure}[!ht]
\captionsetup[subfigure]{labelformat=empty}
\centering
\subfloat[Reference]{\includegraphics[scale = 0.2]{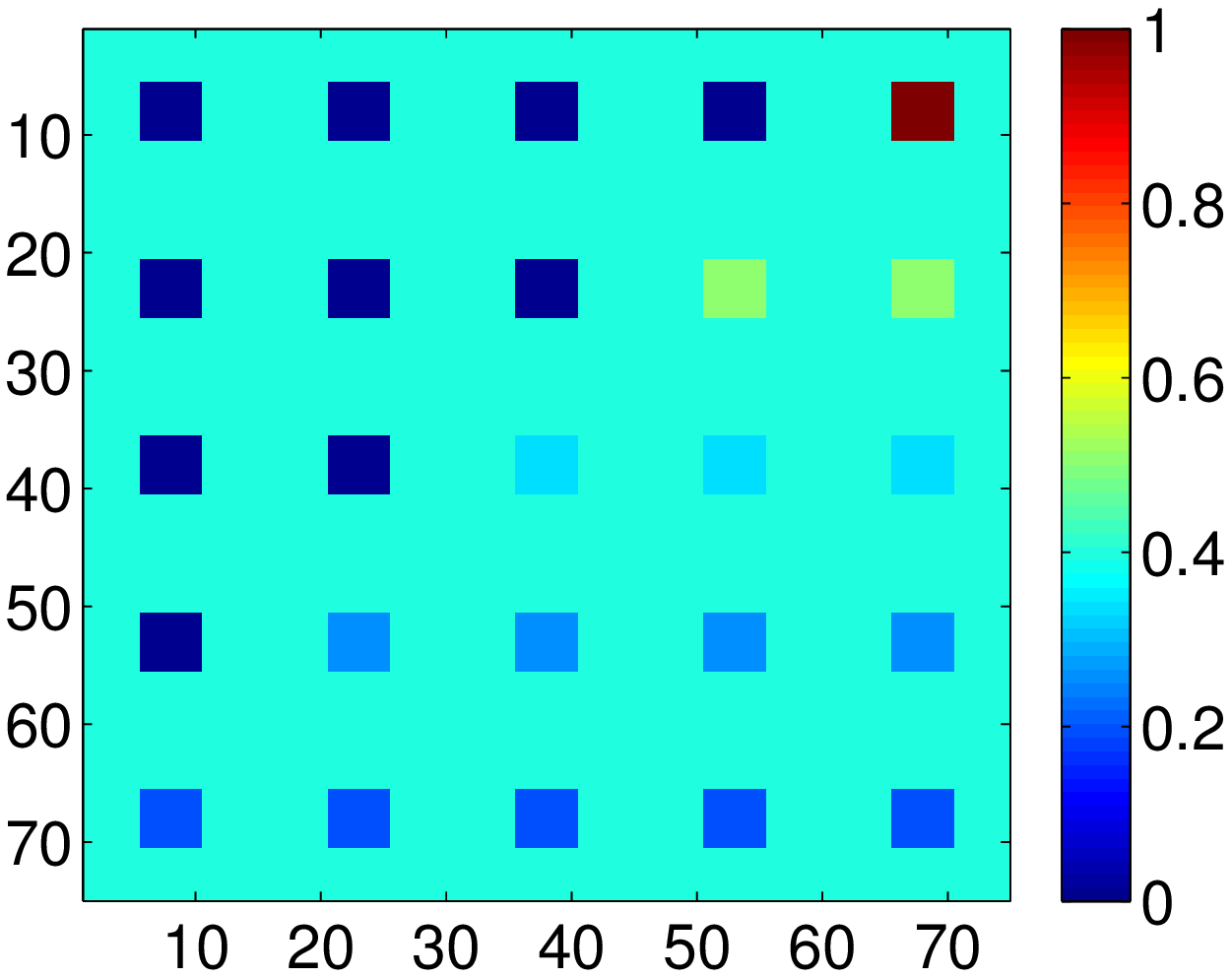}     \label{b5}}
\subfloat[SUnSAL-TV]{\includegraphics[scale = 0.2]{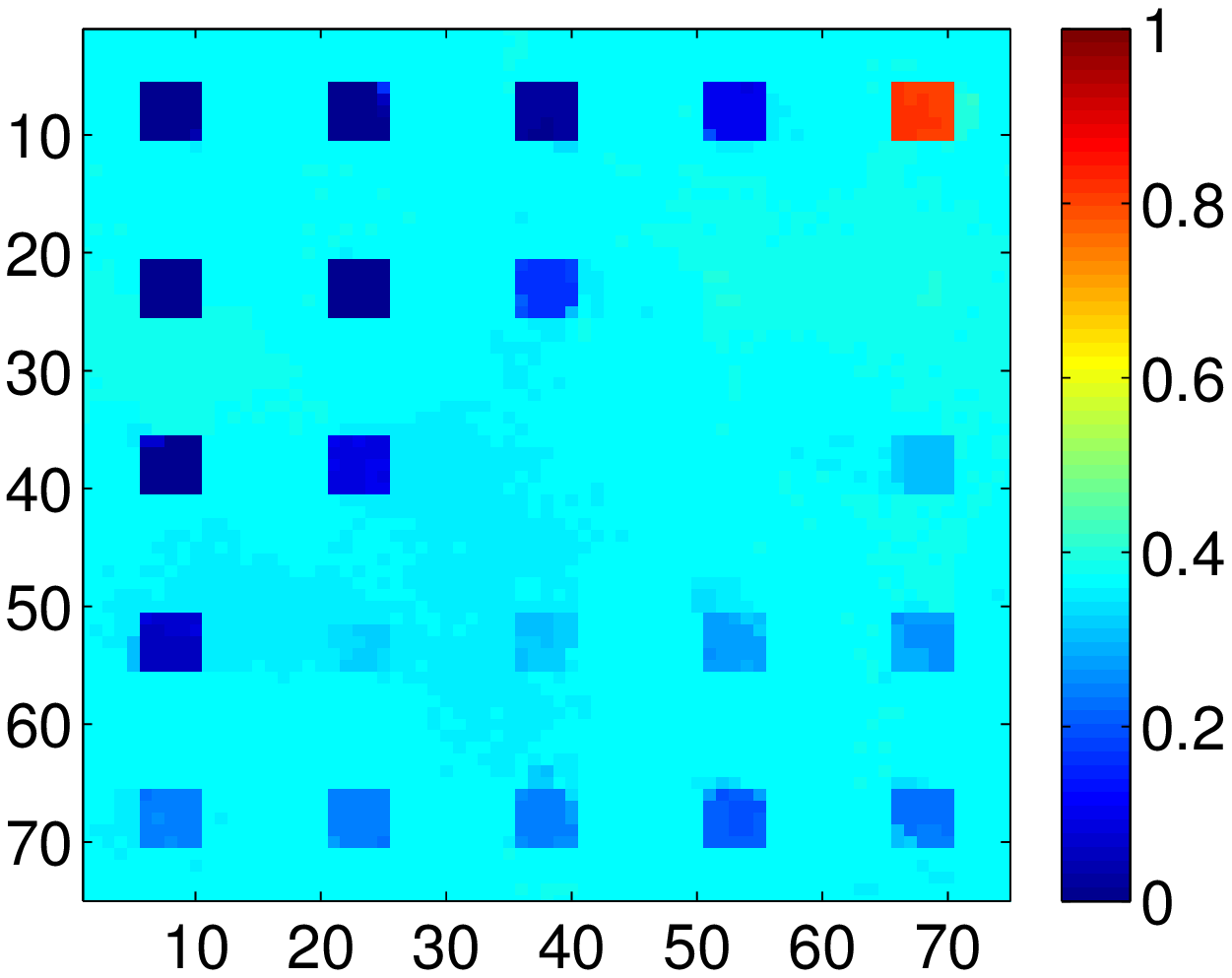}   \label{sunsaltvb5}}
\subfloat[$\textmd{S}^{2}$WSU ]{\includegraphics[scale = 0.2]{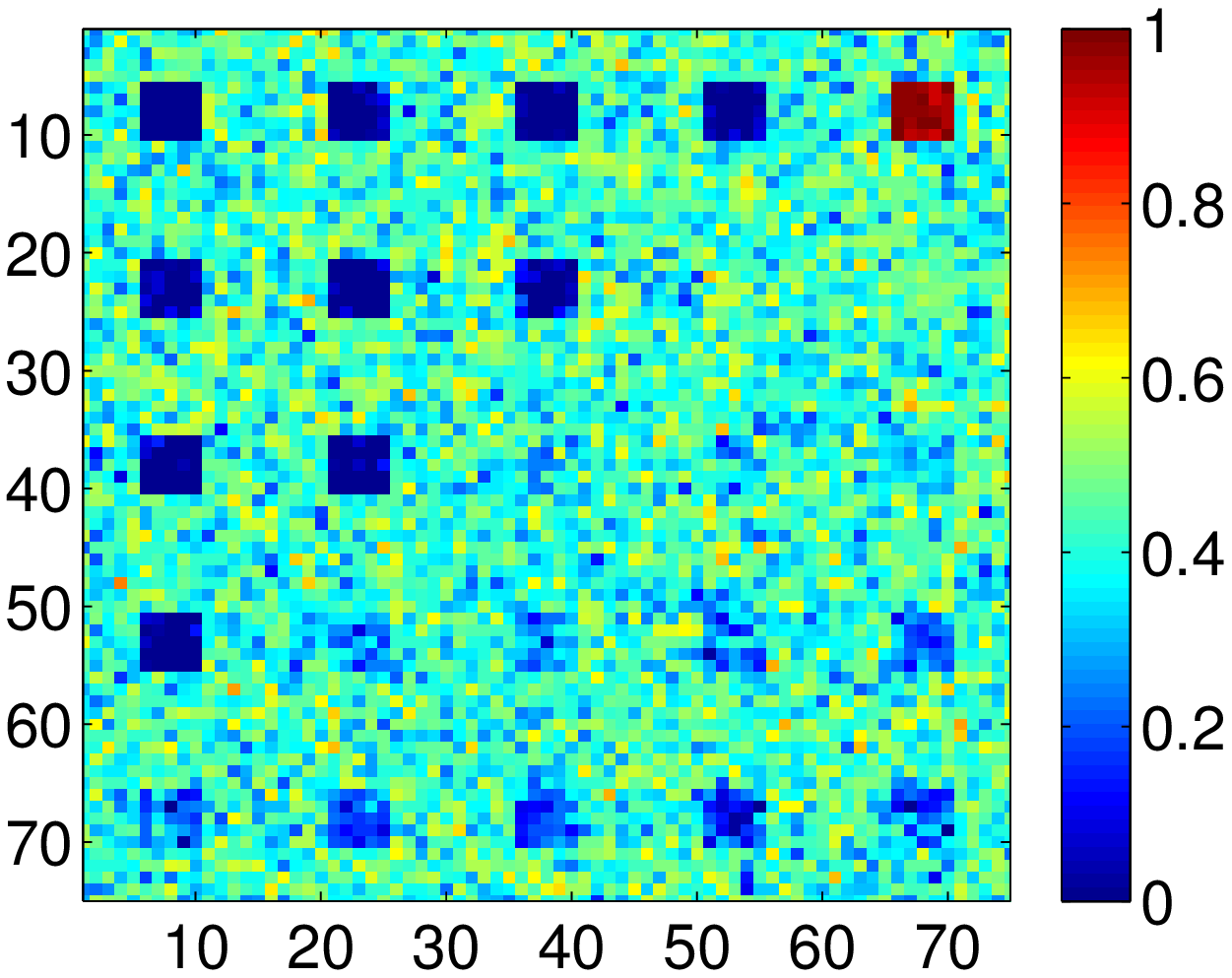}   \label{s2wsub5}}

\subfloat[MUA$_{\textmd{SLIC}}$]{\includegraphics[scale = 0.2]{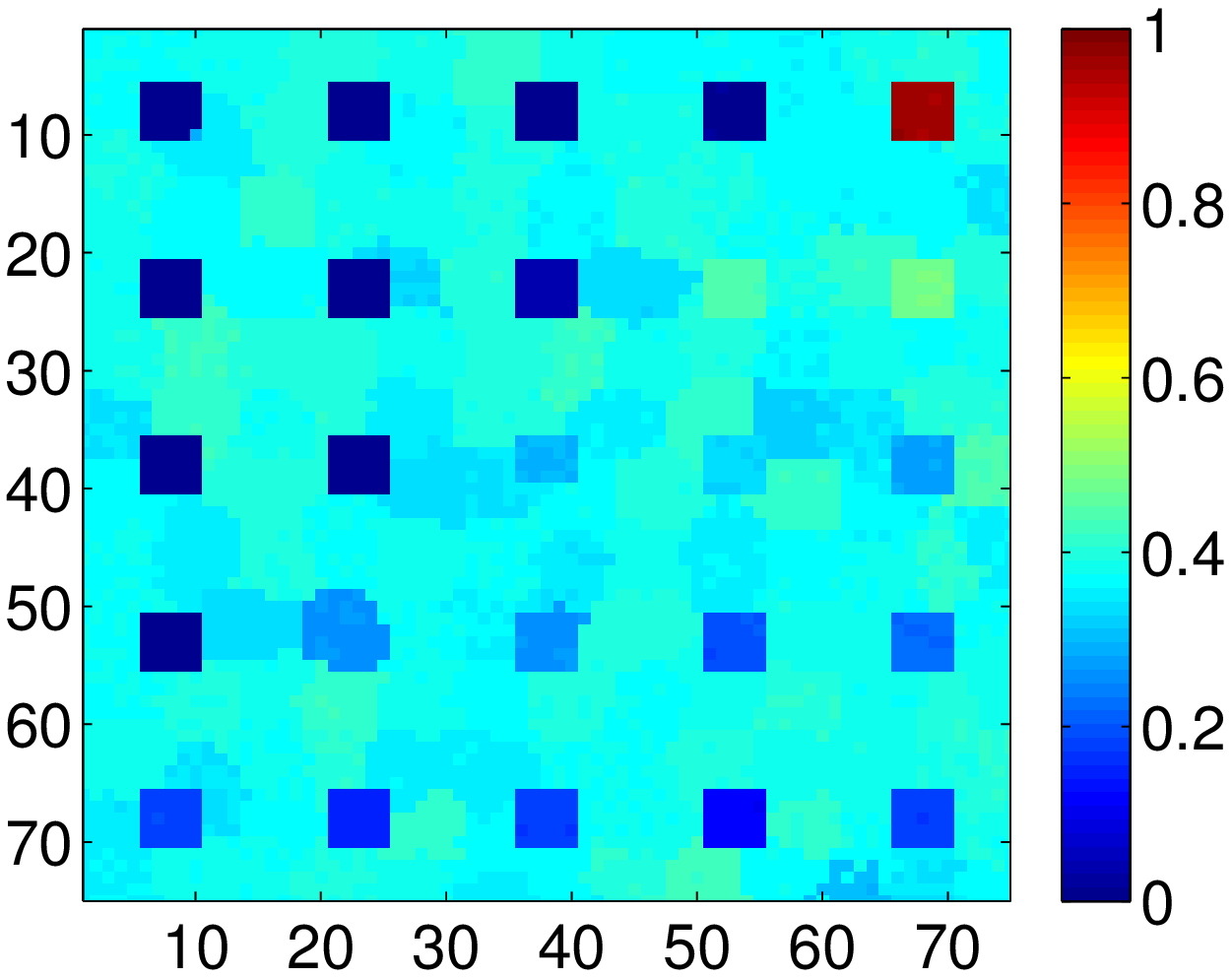}   \label{muaslicb5}}
\subfloat[SUSRLR-TV]{\includegraphics[scale = 0.2]{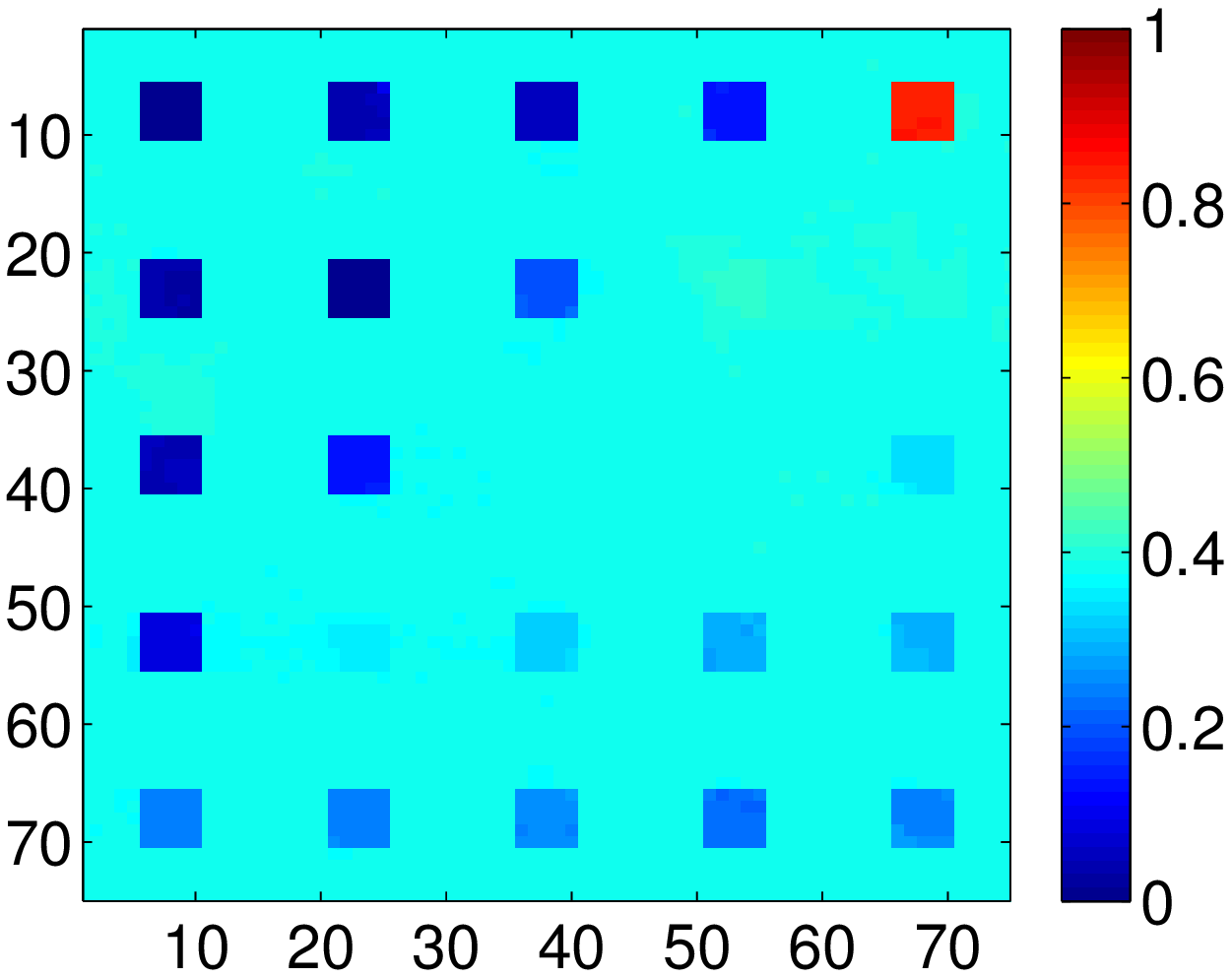}   \label{susrlrtvb5}}
\subfloat[SBGLSU]{\includegraphics[scale = 0.2]{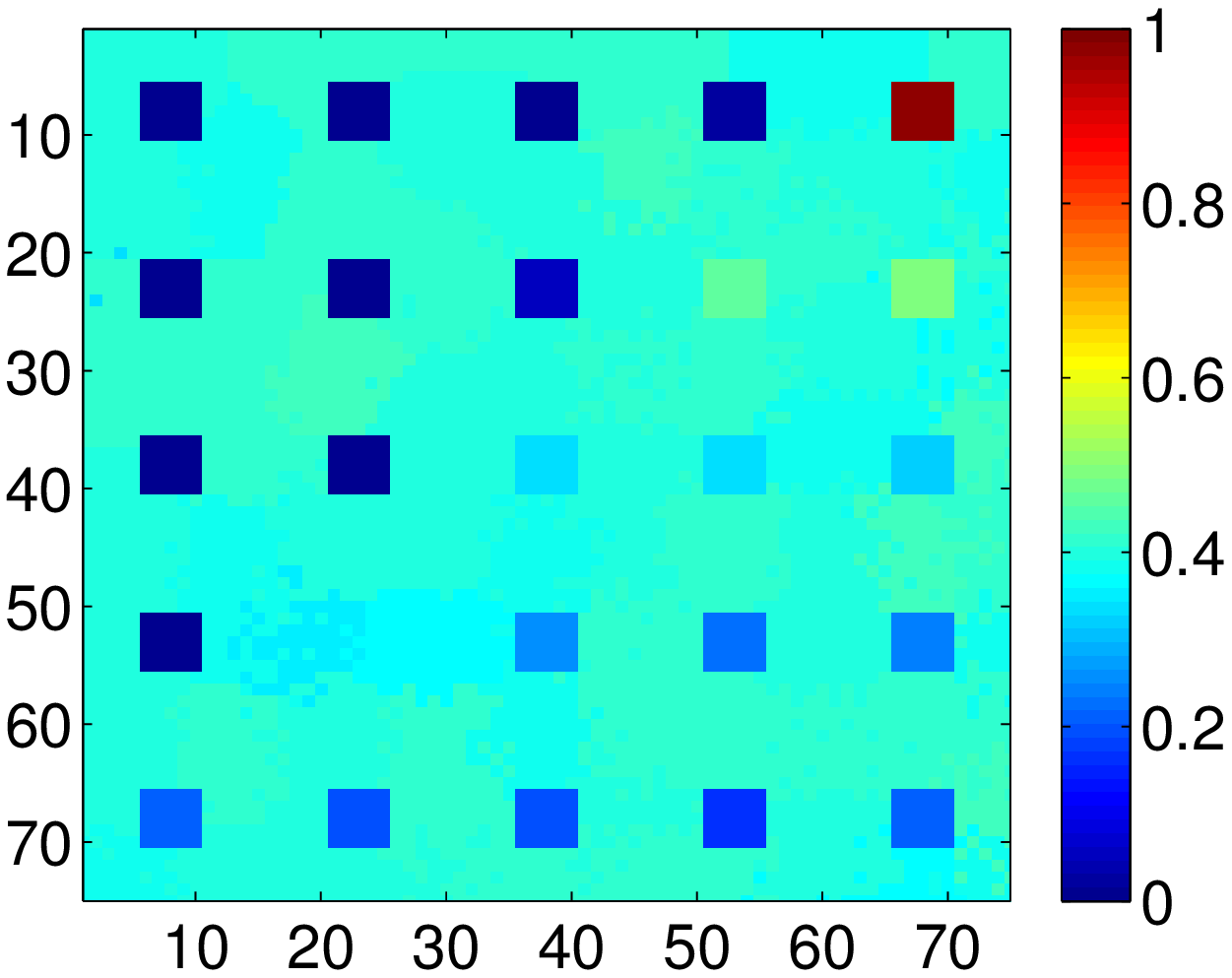}   \label{sggraphb5}}
\caption{Estimated abundance maps for endmember \#5 in DC1 with SNR = 20 dB.}
\label{figdc1}
\end{figure}

\begin{figure}[!ht]
\captionsetup[subfigure]{labelformat=empty}
\centering
\subfloat[Reference]{\includegraphics[scale = 0.2]{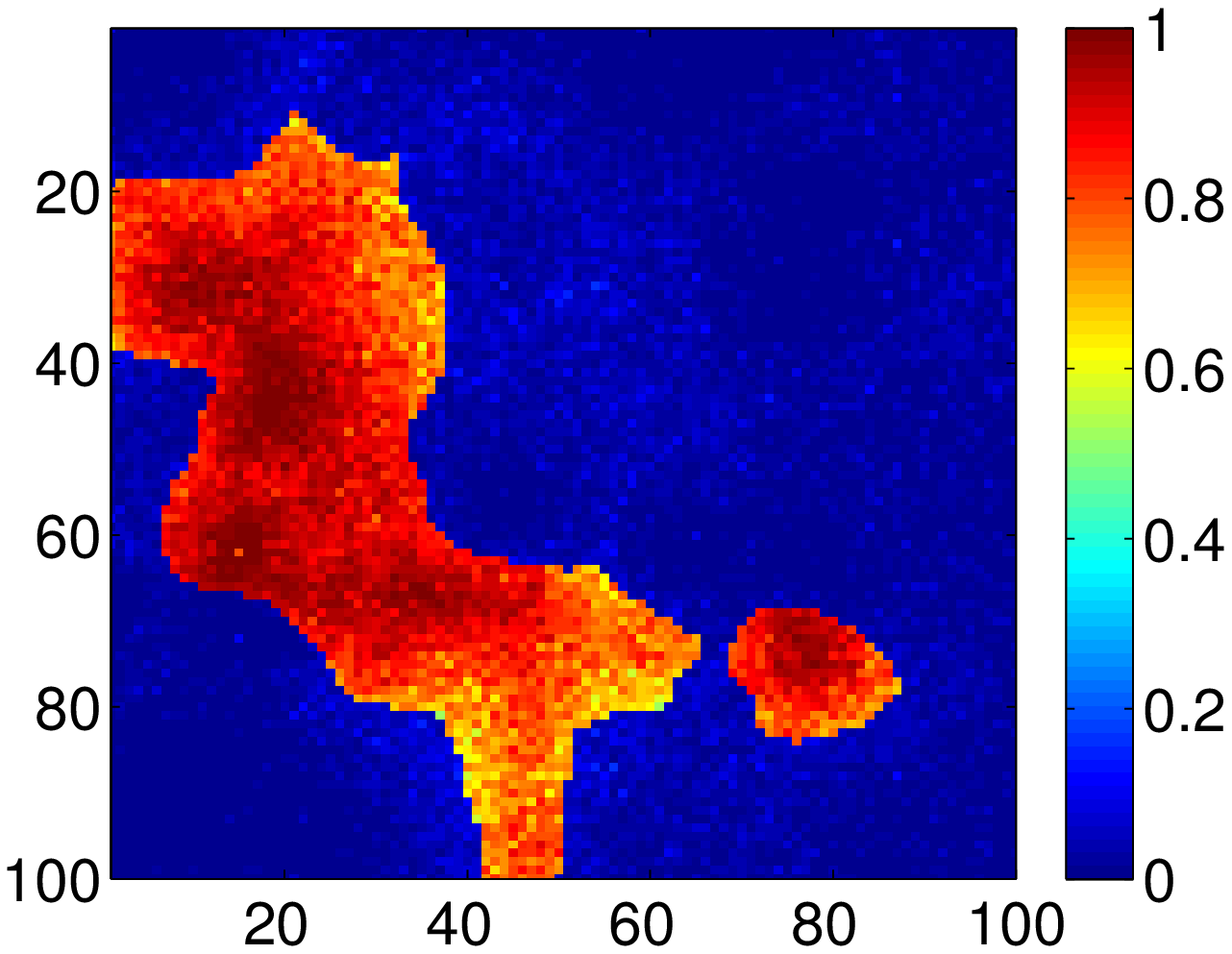}     \label{b1}}
\subfloat[SUnSAL-TV]{\includegraphics[scale = 0.2]{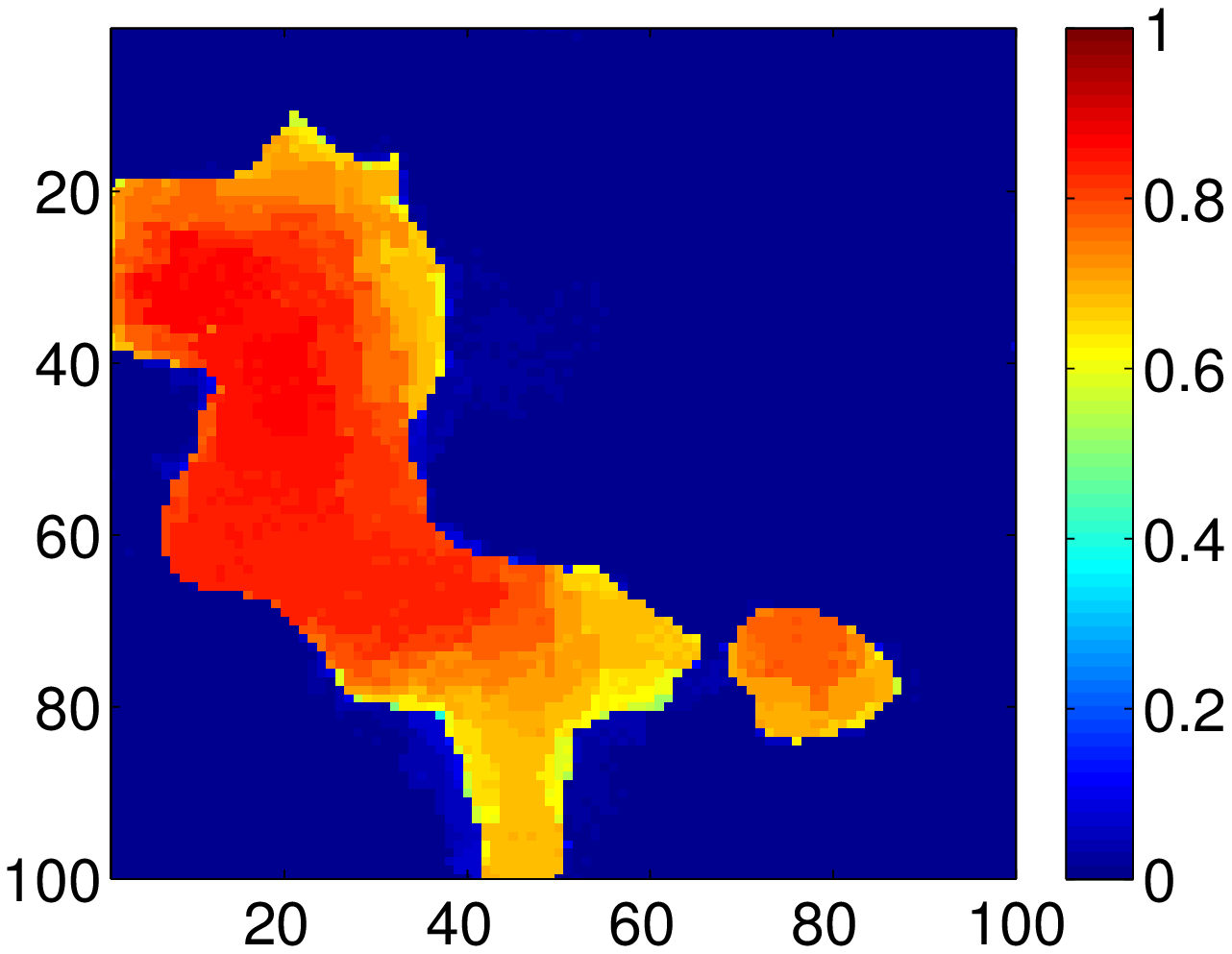}   \label{sunsaltvb1}}
\subfloat[$\textmd{S}^{2}$WSU ]{\includegraphics[scale = 0.2]{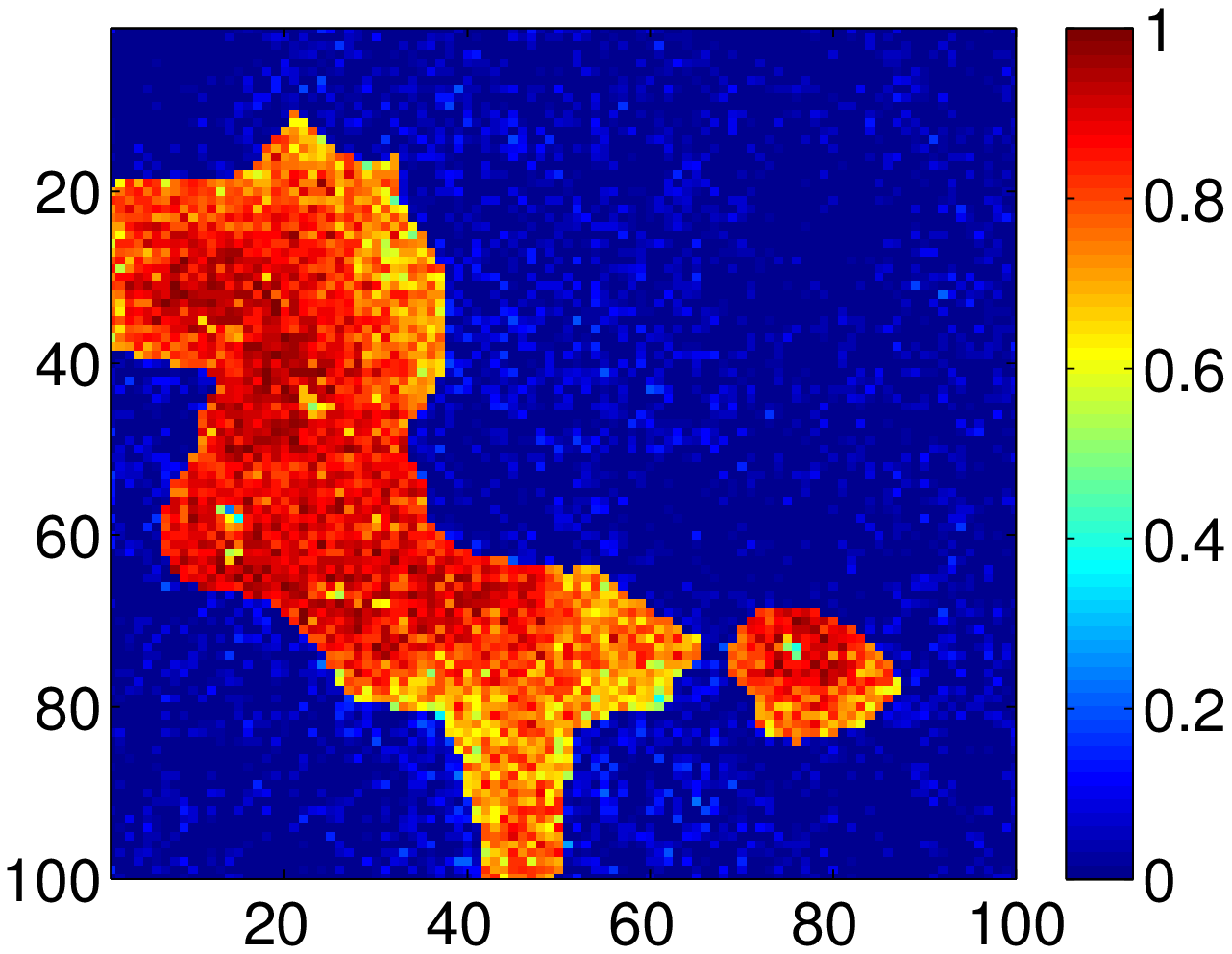}   \label{s2wsub1}}

\subfloat[MUA$_{\textmd{SLIC}}$]{\includegraphics[scale = 0.2]{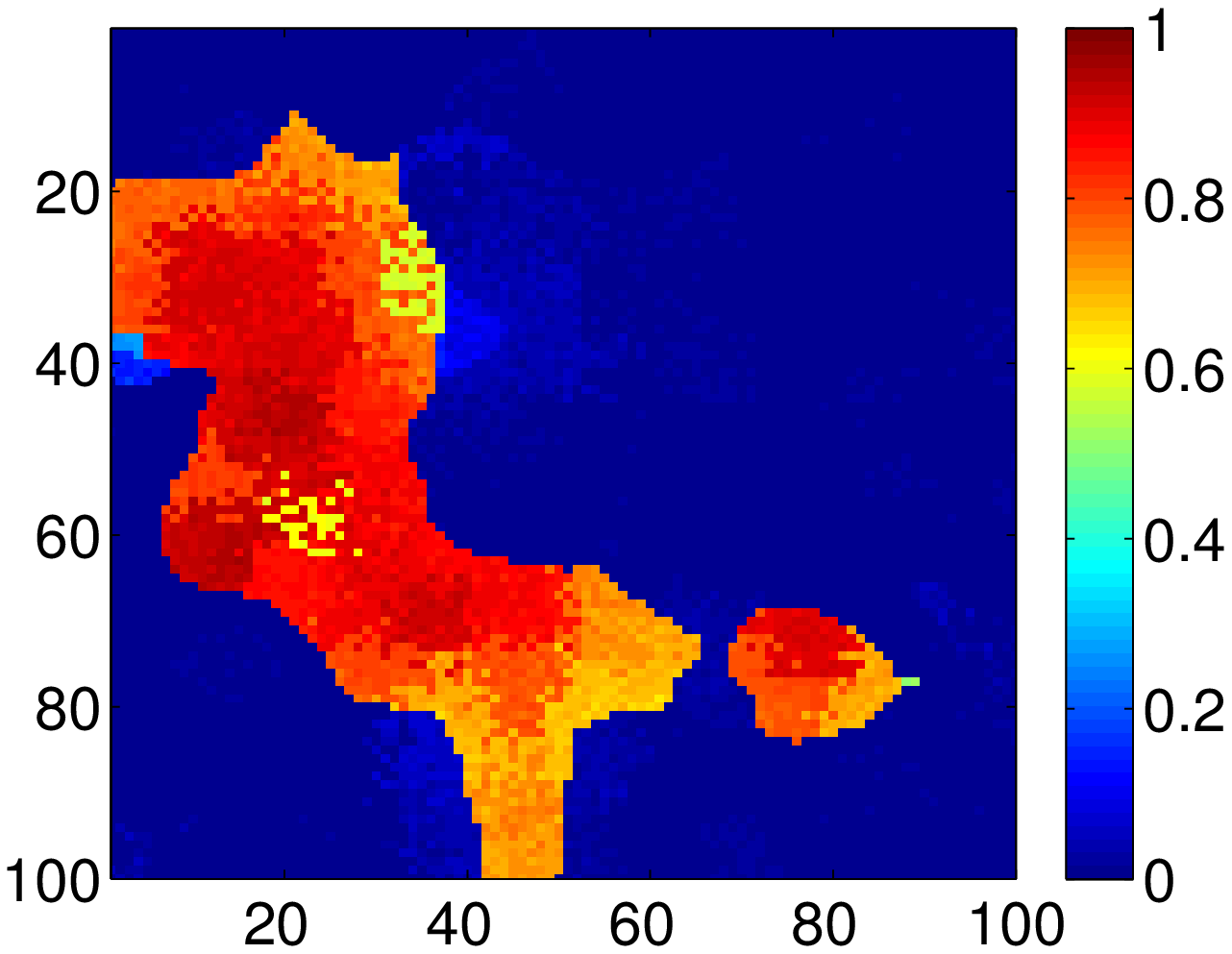}   \label{muaslicb1}}
\subfloat[SUSRLR-TV]{\includegraphics[scale = 0.2]{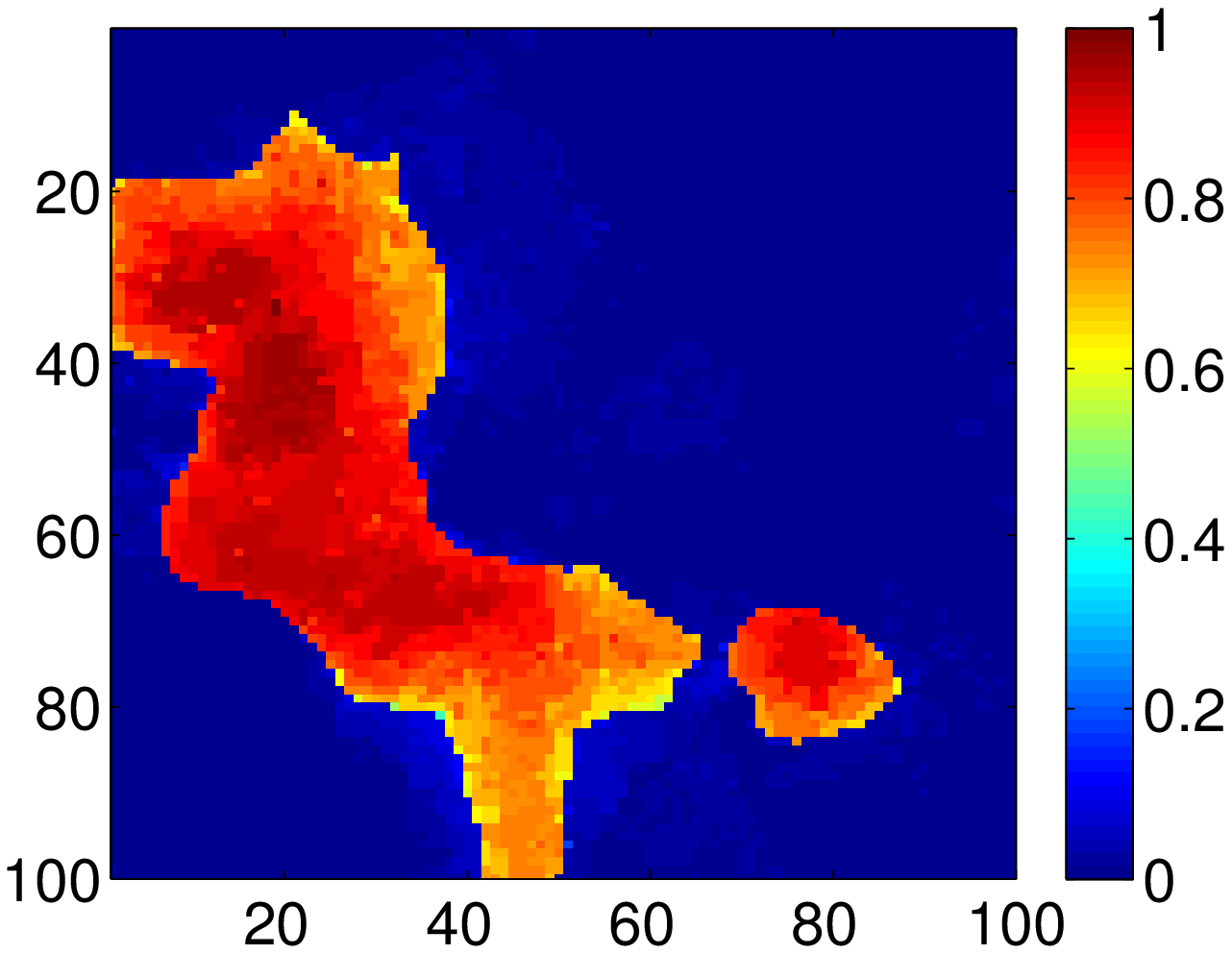}   \label{susrlrtvb1}}
\subfloat[SBGLSU]{\includegraphics[scale = 0.2]{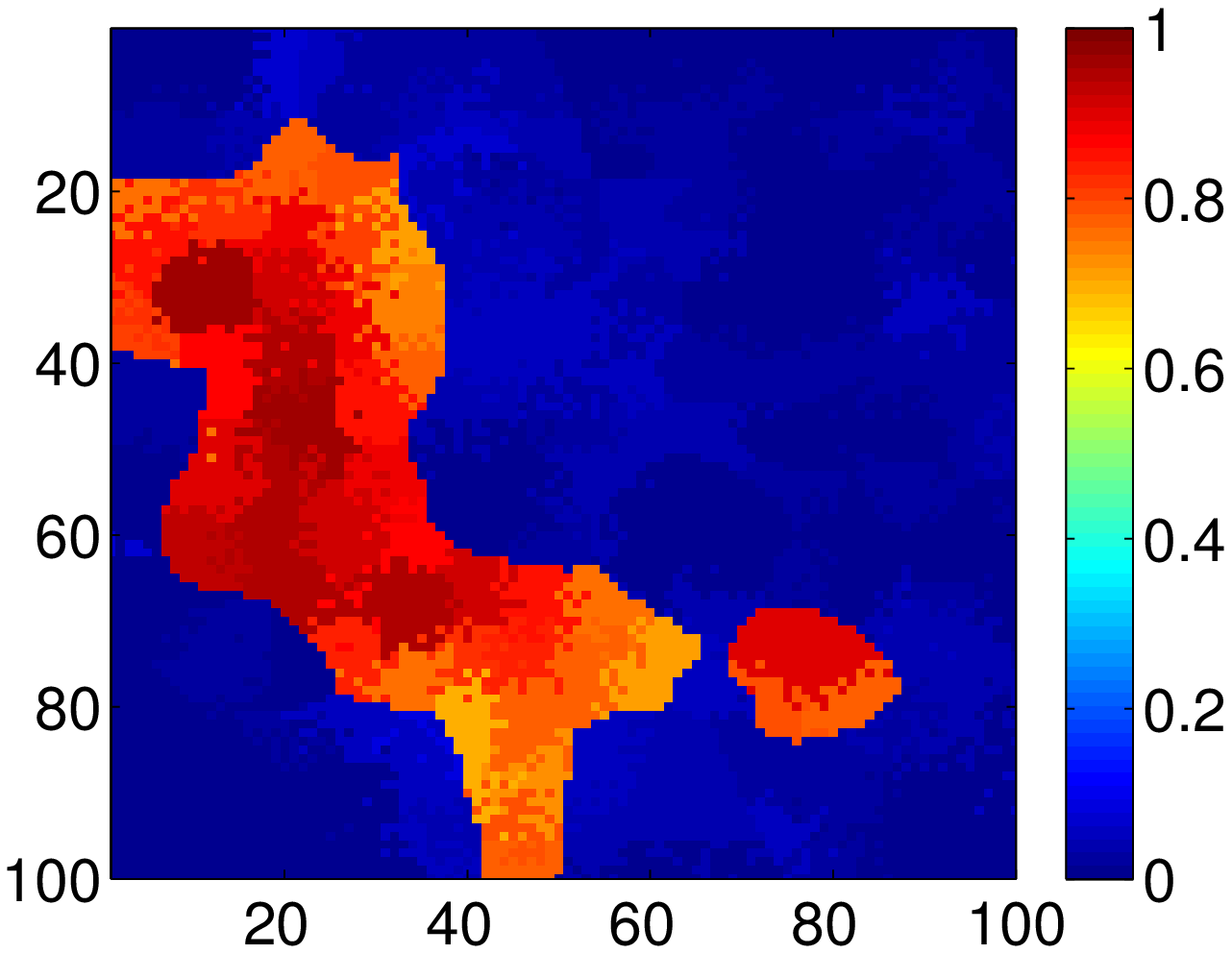}   \label{sggraphb1}}
\caption{Estimated abundance maps for endmember \#1 in DC2 with SNR = 20 dB.}
\label{figdc2}
\end{figure}

We compare the unmixing results of SBGLSU with SUnSAL-TV \cite{6196219}, $\textmd{S}^{2}$WSU \cite{8288814}, MUA$_{\textmd{SLIC}}$ \cite{8536490} and SUSRLR-TV \cite{9099278}. Optimal regularization parameters of all algorithms are found by varying the regularization parameters in a suitable range. The superpixel size and regularization parameter of SLIC algorithm for SBGLSU are set to 8 and 2e-3 for DC1 and DC2, respectively. We run all algorithms under comparison and report the SRE values in Table \ref{table1} along with optimal regularization parameters obtained by different algorithms for DC1 and DC2 for SNR values 20, 30 and 40 dB. We can see clearly that the SBGLSU performs best in all noise levels for DC1 and DC2. SUnSAL-TV has lowest SRE values in all SNR values so it is not reported in Table \ref{table1}.
\begin{table}[!ht]
\renewcommand{\arraystretch}{1.3}
\caption{SRE VALUES OF DIFFERENT ALGORITHMS}
\label{table1}
\centering
\scriptsize
\begin{tabular}{c c c c c c}
\hline\hline
\multicolumn{5}{c}{\textbf{DC1}} \\
\hline
 \textbf{SNR} &  $\textmd{\textbf{S}}^{\textbf{2}}$\textbf{WSU} & \textbf{MUA}$_{\textmd{\textbf{SLIC}}}$ & \textbf{SUSRLR-TV} &\textbf{SBGLSU} \\
\hline
\multirow{1}{*}{\textbf{20}}  &
    \begin{tabular}[x]{@{}c@{}}7.68\\$\lambda=1e-1$\\$$\end{tabular}&
   \begin{tabular}[x]{@{}c@{}}11.34\\$\lambda_{1}=3e-2$\\$\lambda_{2}=1e-1$\end{tabular}&
   \begin{tabular}[x]{@{}c@{}}14.38\\$\rho=1e-1$\\$\lambda_{TV}=5e-2$\end{tabular}&
   \begin{tabular}[x]{@{}c@{}}\textbf{ 19.99 }\\$\lambda_{s}=5e-2$\\$\lambda_{g}=1e3$\end{tabular} \\
\hline
\multirow{1}{*}{\textbf{30}}  &
    \begin{tabular}[x]{@{}c@{}}15.48\\$\lambda=5e-3$\\$$\end{tabular}&
   \begin{tabular}[x]{@{}c@{}}15.73\\$\lambda_{1}=7e-3$\\$\lambda_{2}=5e-2$\end{tabular}&
   \begin{tabular}[x]{@{}c@{}}25.29\\$\rho=5e-2$\\$\lambda_{TV}=1e-2$\end{tabular}&
   \begin{tabular}[x]{@{}c@{}}\textbf{ 34.49 }\\$\lambda_{s}=1e-2$\\$\lambda_{g}=1e3$\end{tabular} \\
\hline
\multirow{1}{*}{\textbf{40}}  &
    \begin{tabular}[x]{@{}c@{}}28.23\\$\lambda=1e-3$\\$$\end{tabular}&
   \begin{tabular}[x]{@{}c@{}}22.34\\$\lambda_{1}=1e-3$\\$\lambda_{2}=1e-2$\end{tabular}&
   \begin{tabular}[x]{@{}c@{}}38.72\\$\rho=1e-2$\\$\lambda_{TV}=5e-4$\end{tabular}&
   \begin{tabular}[x]{@{}c@{}}\textbf{ 45.33 }\\$\lambda_{s}=5e-3$\\$\lambda_{g}=1e3$\end{tabular} \\
\hline\hline

\multicolumn{5}{c}{\textbf{DC2}} \\
\hline
 \textbf{SNR} &  $\textmd{\textbf{S}}^{\textbf{2}}$\textbf{WSU} & \textbf{MUA}$_{\textmd{\textbf{SLIC}}}$ & \textbf{SUSRLR-TV} &\textbf{SBGLSU}\\
\hline
\multirow{1}{*}{\textbf{20}}  &
    \begin{tabular}[x]{@{}c@{}}9.33\\$\lambda=1e-1$\\$$\end{tabular}&
   \begin{tabular}[x]{@{}c@{}}14.75\\$\lambda_{1}=3e-2$\\$\lambda_{2}=1e-1$\end{tabular}&
   \begin{tabular}[x]{@{}c@{}}16.08\\$\rho=1e-1$\\$\lambda_{TV}=5e-2$\end{tabular}&
   \begin{tabular}[x]{@{}c@{}}\textbf{ 18.13 }\\$\lambda_{s}=2e-2$\\$\lambda_{g}=1e3$\end{tabular} \\
\hline
\multirow{1}{*}{\textbf{30}}  &
    \begin{tabular}[x]{@{}c@{}}21.66\\$\lambda=5e-3$\\$$\end{tabular}&
   \begin{tabular}[x]{@{}c@{}}18.33\\$\lambda_{1}=7e-3$\\$\lambda_{2}=5e-2$\end{tabular}&
   \begin{tabular}[x]{@{}c@{}}22.25\\$\rho=5e-2$\\$\lambda_{TV}=1e-2$\end{tabular}&
   \begin{tabular}[x]{@{}c@{}}\textbf{ 23.51 }\\$\lambda_{s}=7e-2$\\$\lambda_{g}=5e-2$\end{tabular} \\
\hline
\multirow{1}{*}{\textbf{40}}  &
    \begin{tabular}[x]{@{}c@{}}27.79\\$\lambda=1e-3$\\$$\end{tabular}&
   \begin{tabular}[x]{@{}c@{}}20.92\\$\lambda_{1}=1e-3$\\$\lambda_{2}=5e-3$\end{tabular}&
   \begin{tabular}[x]{@{}c@{}}25.97\\$\rho=5e-3$\\$\lambda_{TV}=1e-3$\end{tabular}&
   \begin{tabular}[x]{@{}c@{}}\textbf{ 29.52 }\\$\lambda_{s}=2e-2$\\$\lambda_{g}=7e-3$\end{tabular} \\
   \hline\hline
\end{tabular}
\end{table}
Furthermore, we compare the unmixing results visually for individual endmember. Fig. \ref{figdc1} shows the estimated abundance map obtained by different unmixing algorithms for endmember \#5 in DC1 with SNR = 20 dB. It can be seen clearly that SBGLSU is able to recover the details much better than the other algorithms. Similar conclusions can be made for DC2. Fig. \ref{figdc2} shows the estimated abundance map obtained by different unmixing algorithms for endmember \#1 in DC2 with SNR = 20 dB.

\begin{figure}[!ht]
\centering
\subfloat[DC1]{\includegraphics[scale=0.32]{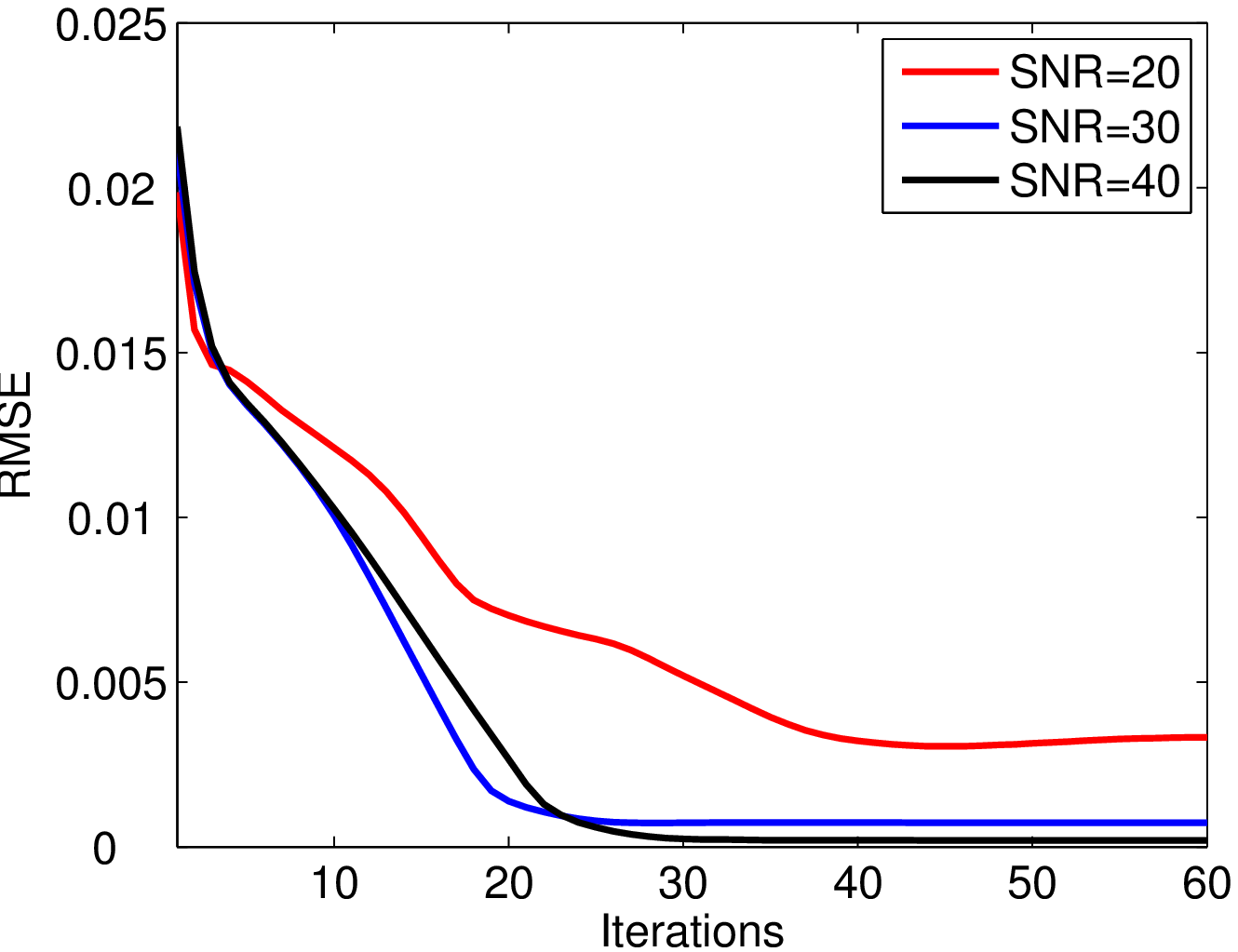}
\label{fig3a}}
\subfloat[DC2]{\includegraphics[scale=0.32]{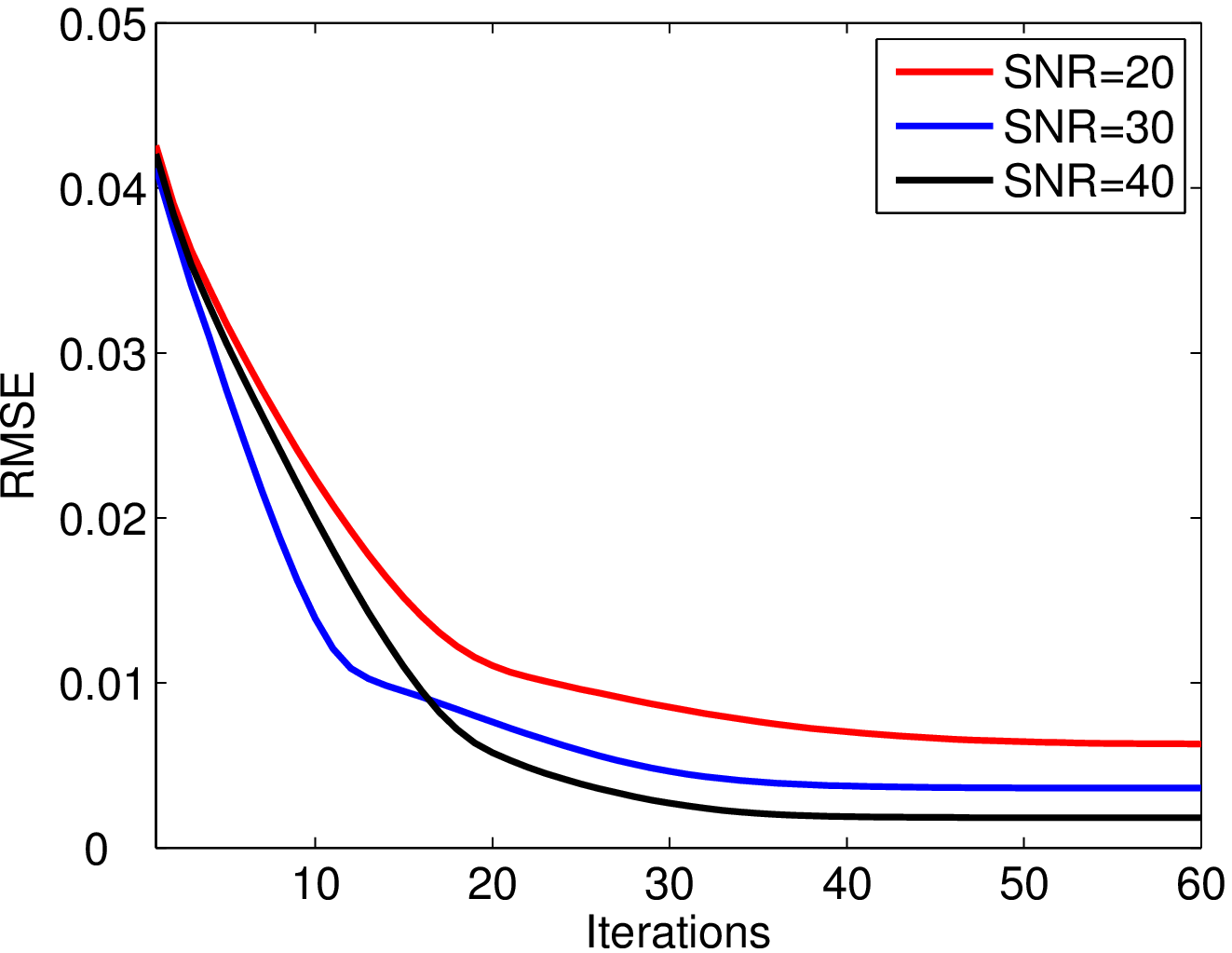}
\label{fig3b}}
\caption{Convergence curves of SBGLSU.}
\label{conver}
\end{figure}

For convergency analysis, we record the root mean square error (RMSE) at each outer iteration of SBGLSU. RMSE is defined as $\textmd{RMSE} = \sqrt{\frac{1}{mn}\sum_{i=1}^{n}\|\mathbf{s}_{i}-\hat{\mathbf{{s}}}_{i}\|^{2}}$ where $\mathbf{s}_{i}$ and $\mathbf{\hat{s}}_{i}$ are the actual and estimated abundance vectors, respectively. Fig. \ref{conver} shows the convergency curve for SBGLSU for DC1 and DC2 at all noise levels. We can see that after 60 outer iterations, SBGLSU is able to obtain stable solution.
\subsection{Real Data Experiment}
In this section, we present the performance of the proposed method on the real data set. The real data set used in the experiment is Cuprite dataset\footnote{http://aviris.jpl.nasa.gov/html/aviris.freedata.html} which contains 14 kinds of minerals \cite{1411995}. This data set is used frequently to compare the performance of the unmixing algorithms. It contains 224 bands with a wavelength range of 0.4-2.5 $\mu$m. However, some bands of the Cuprite data set have low-SNR and water absorption. Therefore, we removed the bands 1-2, 105-115, 150-170, and 223-224 prior to analysis. The spatial size of the data used in the experiment is $250\times191$. We use the spectral library of 498 minerals from the USGS library.
Since we do not have exact abundance maps for Cuprite data, we use Tetracorder 4.4 \cite{Clark2003Imaging} classification algorithm in order to compare the unmixing results qualitatively. We compare the unmixing results of SBGLSU with SUnSAL-TV, $\textmd{S}^{2}$WSU, MUA$_{\textmd{SLIC}}$, SUSRLR-TV for Chalcedony mineral in the Cuprite data. The regularization parameters of the algorithms under comparison are set to as: $\lambda = 10^{-3}$ and $\lambda_{TV} = 10^{-3}$ for SUnSAL-TV, $\lambda = 7\times10^{-1}$ for $\textmd{S}^{2}$WSU, $\lambda_{1} = 10^{-3}$ and $\lambda_{2} = 10^{-3}$ for MUA$_{\textmd{SLIC}}$, $\rho = 10^{-3}$ and $\lambda_{TV} = 10^{-3}$ for SUSRLR-TV and $\lambda_{s} = 10^{-3}$ and $\lambda_{g} = 10^{-3}$ for SBGLSU.

\begin{figure}[!ht]
\centering
\captionsetup[subfigure]{labelformat=empty}
\subfloat[Tetracorder 4.4]{\includegraphics[width=2cm,height=2.3cm]{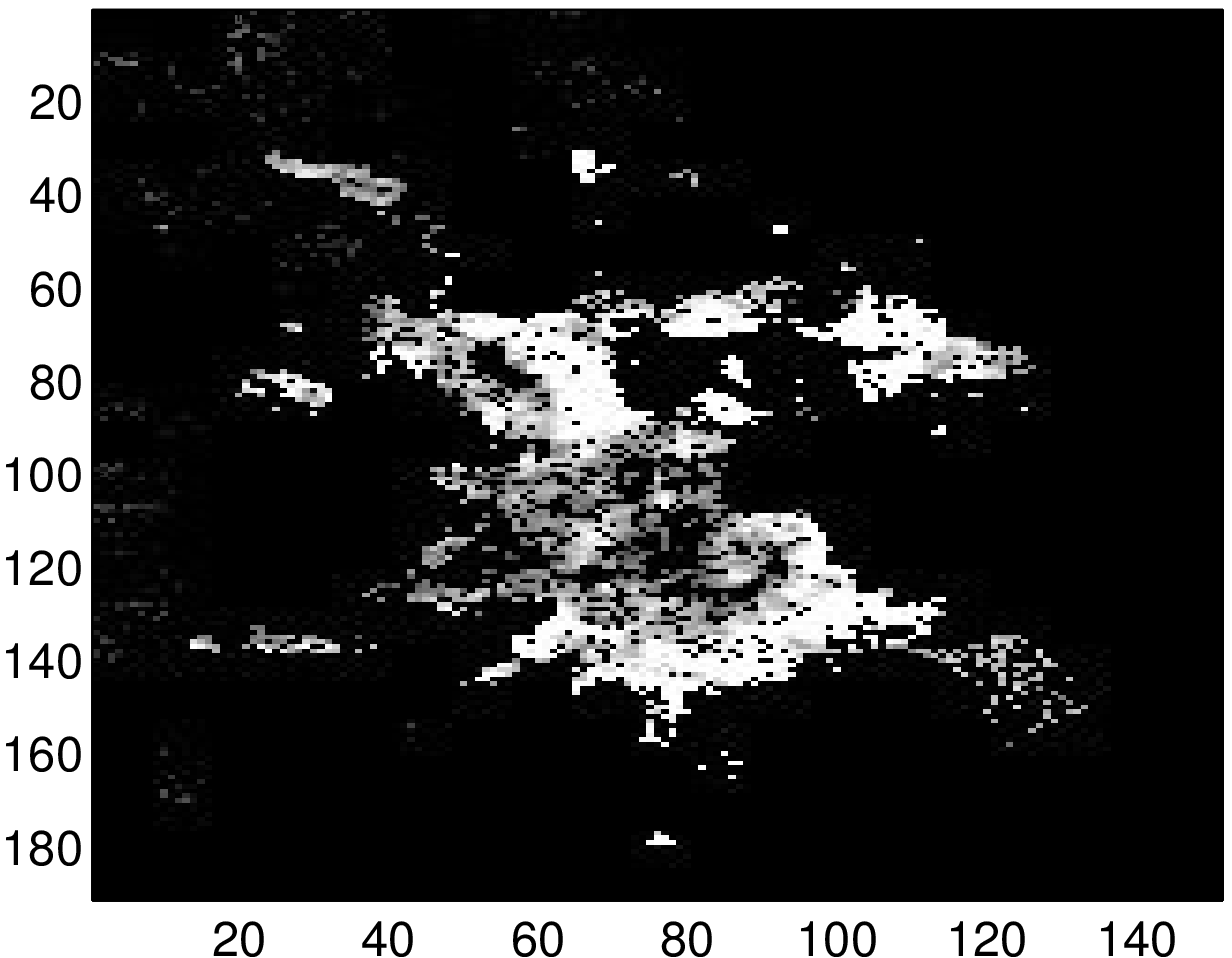}
\label{figm}}
\subfloat[SUnSAL-TV]{\includegraphics[width=2.4cm,height=2.4cm]{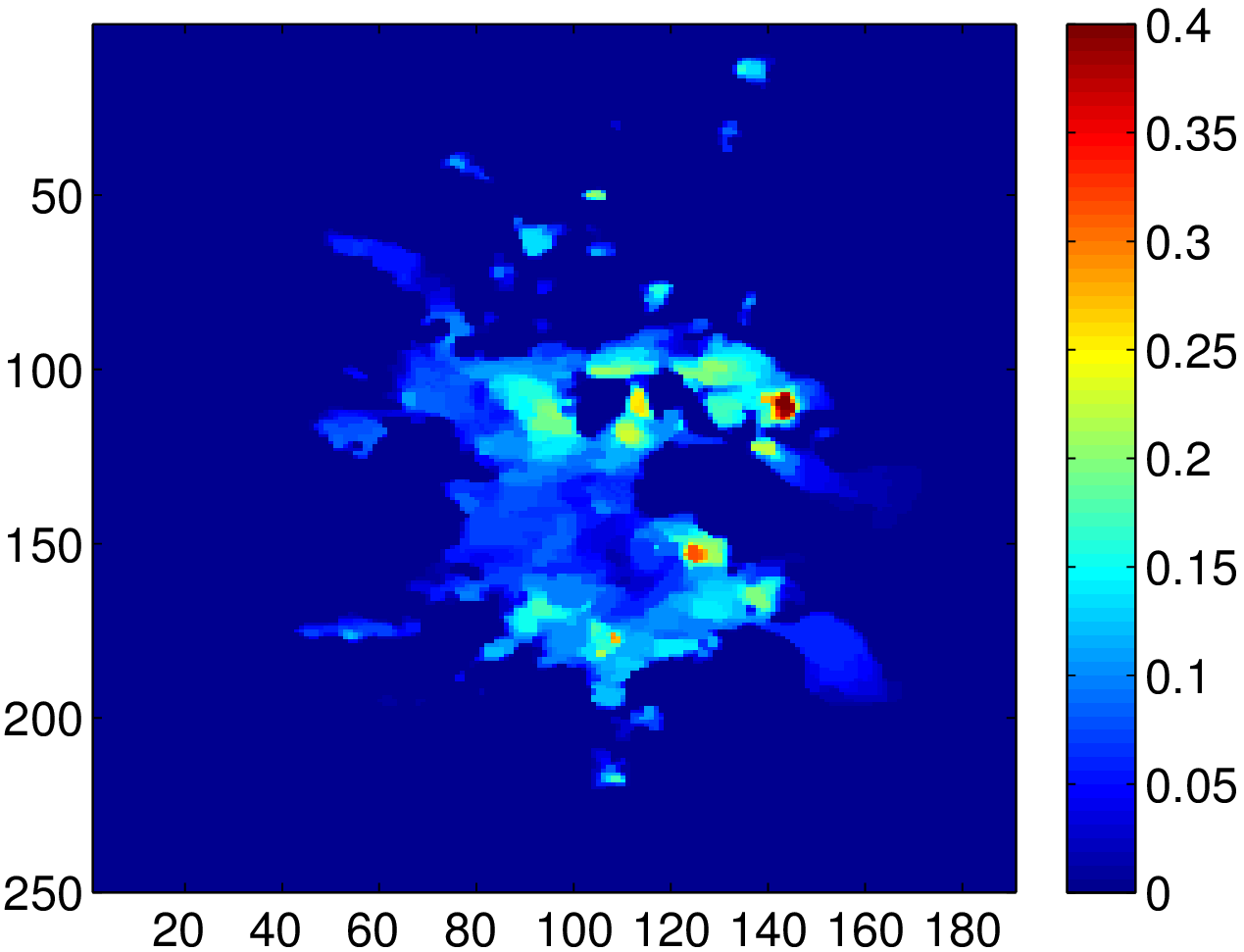}
\label{fign}}
\subfloat[$\textmd{S}^{2}$WSU]{\includegraphics[width=2.4cm,height=2.4cm]{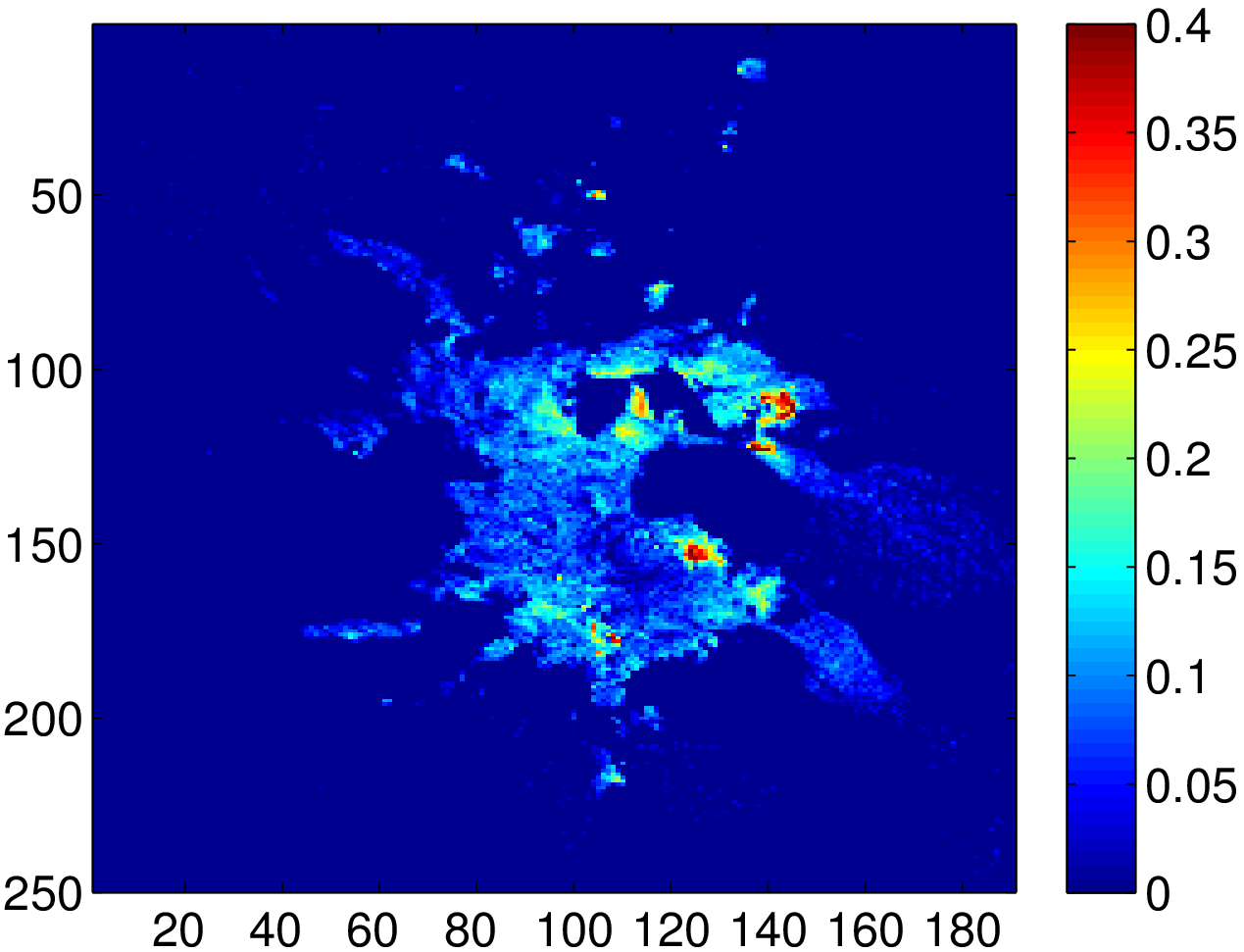}
\label{figo}}

\subfloat[MUA$_{\textmd{SLIC}}$]{\includegraphics[width=2.4cm,height=2.4cm]{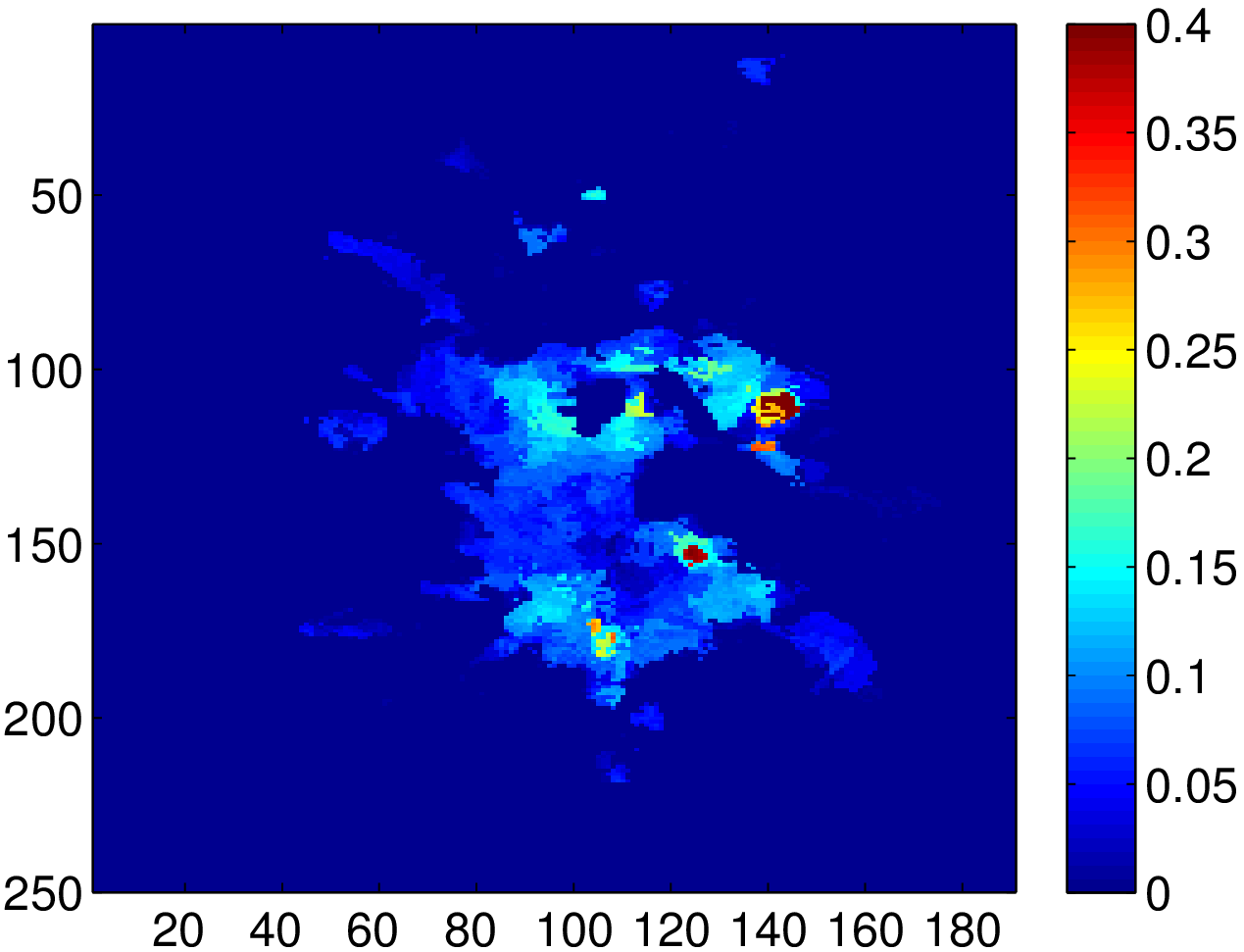}
\label{figp}}
\subfloat[SUSRLR-TV]{\includegraphics[width=2.4cm,height=2.4cm]{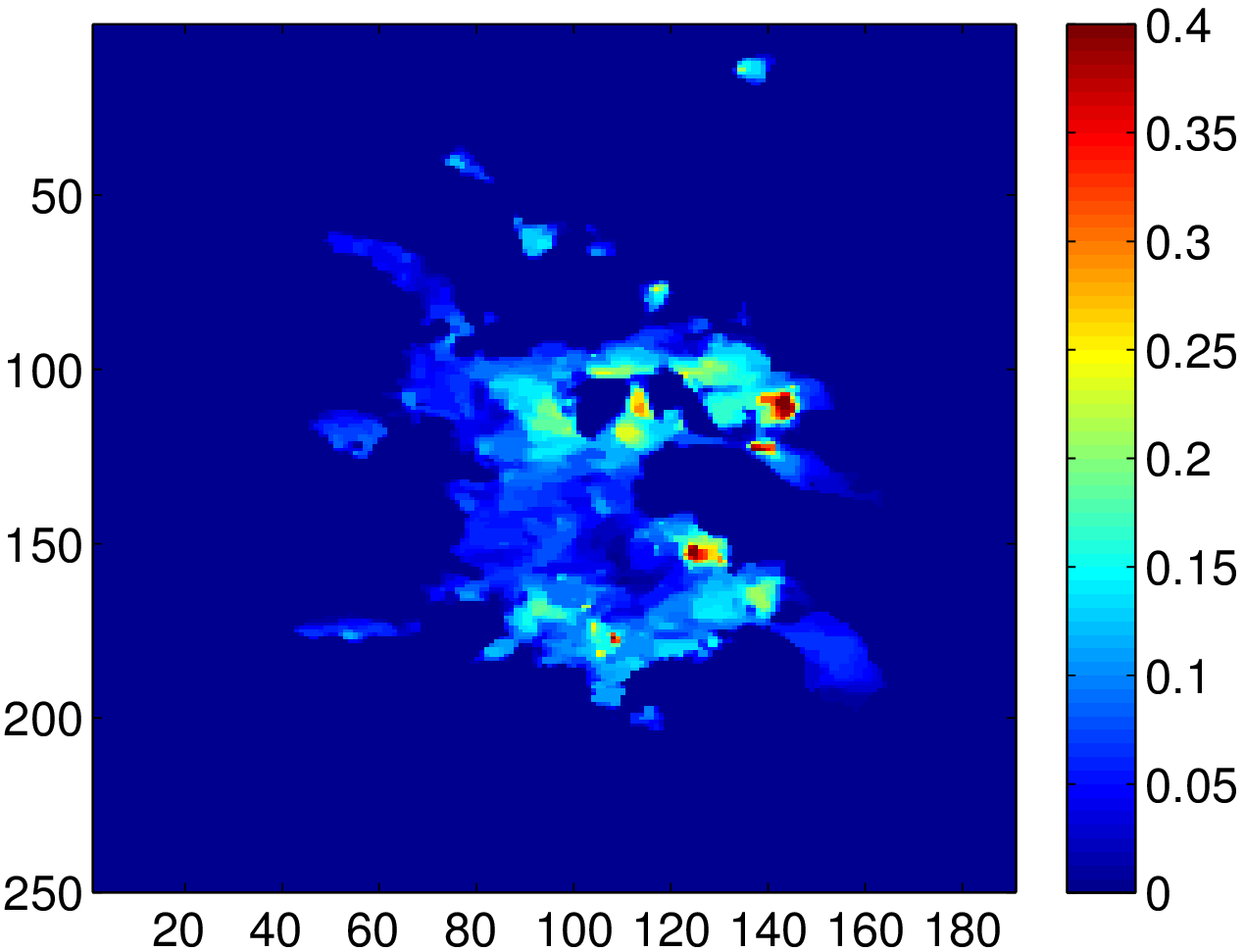}
\label{figq}}
\subfloat[SBGLSU]{\includegraphics[width=2.4cm,height=2.4cm]{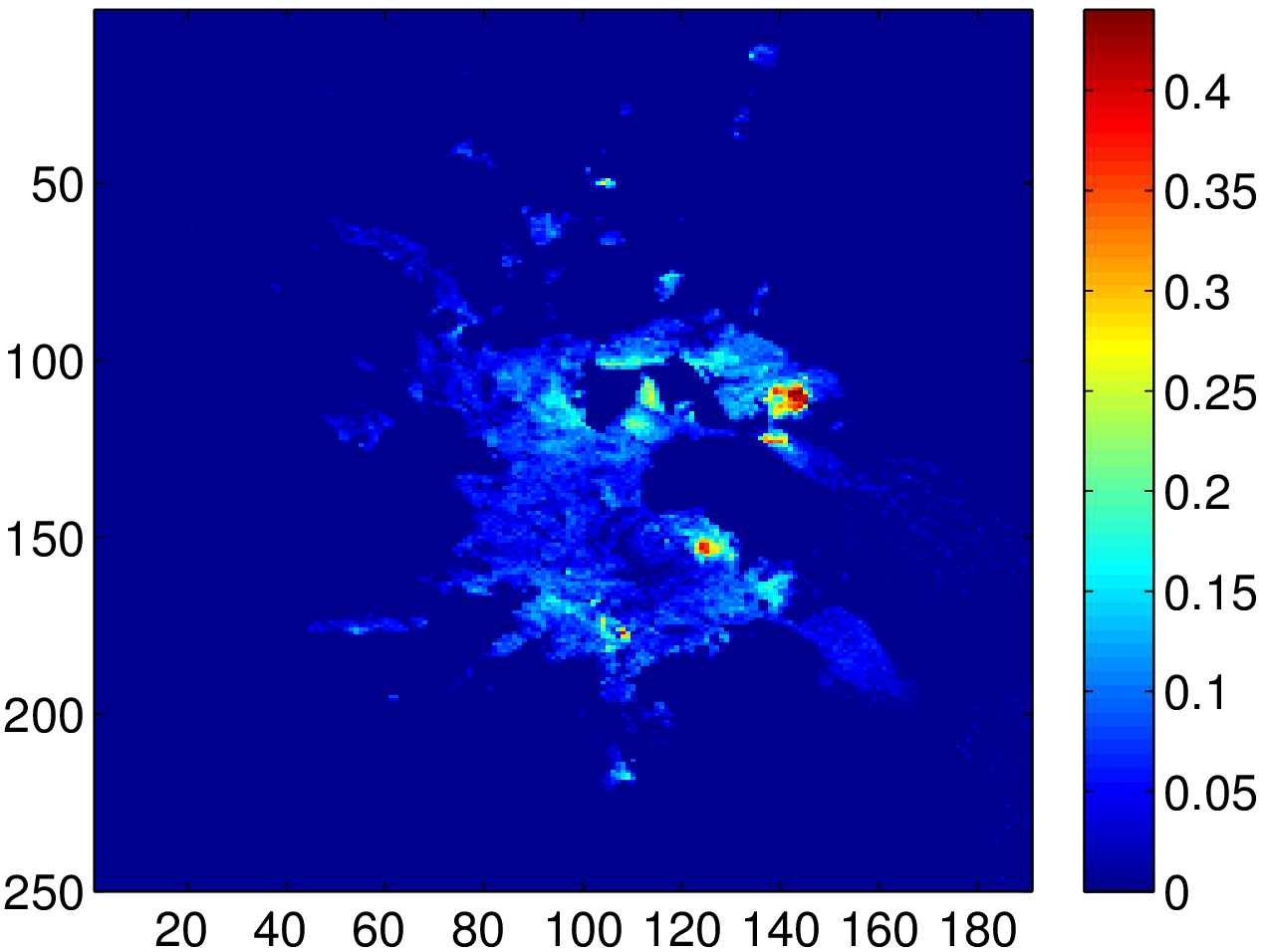}
\label{figr}}
\caption{Abundance maps obtained by different algorithms corresponding Chalcedony.}
\label{figreal}
\end{figure}

Fig. \ref{figreal} shows a qualitative comparison among the classification maps obtained by Tetracorder 4.4 \cite{Clark2003Imaging} algorithm and abundance maps obtained by SBGLSU, SUnSAL-TV, $\textmd{S}^{2}$WSU, MUA$_{\textmd{SLIC}}$ and SUSRLR-TV.  It can be concluded that SBGLSU is a valid unmixing algorithm for real hyperspectral data.

The computation times of all algorithms under comparison are reported in Table \ref{table2}. MUA$_{\textmd{SLIC}}$ is the fastest unmixing algorithm under comparison. SBGLSU computation time is comparably much better than other algorithms except MUA$_{\textmd{SLIC}}$.
\begin{table}[!ht]
\renewcommand{\arraystretch}{1.3}
\caption{COMPUTATION TIMES OF DIFFERENT ALGORITHMS ON REAL DATA (IN MINUTES).}
\label{table2}
\centering
\scriptsize
\begin{tabular}{ c|c|c|c|c}
\hline\hline
\textbf{SUnSAL-TV}  &$\textmd{\textbf{S}}^{\textbf{2}}$\textbf{WSU} & \textbf{MUA}$_{\textmd{\textbf{SLIC}}}$ & \textbf{SUSRLR-TV} &\textbf{SBGLSU}\\
\hline
   28.27&    19.36&   \textbf{2.44}&   50.68&  13.95 \\
\hline\hline
\end{tabular}
\end{table}
\section{Conclusion}
In this paper, we have developed a novel graph Laplacian regularized sparse hyperspectral unmixing method based on superpixel segmentation. Superpixel segmentation extracts the spatially homogeneous regions and graph Laplacian regularization minimizes the abundance similarity of each superpixel. A sparsity inducing norm with a weighting strategy is included in the formulation to promote the sparsity of the abundance matrix better. The proposed method is solved using a variable splitting approach which includes inner and outer loops to converge better. Experimental results on both simulated and real data sets have shown that the proposed method is a very effective sparse unmixing method compared to other state-of-the-art sparse unmixing methods in the literature.
\balance
\bibliographystyle{IEEEtran}
\bibliography{IEEEabrv,refunmix}
\end{document}